\documentclass[11pt]{article}
\usepackage[preprint]{acl}
\usepackage{times}
\usepackage{latexsym}
\usepackage[T1]{fontenc}
\usepackage[utf8]{inputenc}
\usepackage{microtype}
\usepackage{inconsolata}
\usepackage{graphicx}
\usepackage{booktabs}
\usepackage{multirow}
\usepackage{threeparttable}
\usepackage{makecell}
\usepackage{listings}
\usepackage{xspace}
\newcommand{\dataname}{BiomedTLDR\xspace}

\title{Not too long do read: Evaluating LLM-generated extreme scientific summaries}

\author{Zhuoqi Lyu \and Qing Ke \\
        Department of Data Science \\
        City University of Hong Kong \\
        Hong Kong, China \\
   \small{
   \textbf{Correspondence:} \href{mailto:q.ke@cityu.edu.hk}{q.ke@cityu.edu.hk}
 }
}

\begin{document}

\maketitle

\begin{abstract}
High-quality scientific extreme summary (TLDR) facilitates effective science communication. How do large language models (LLMs) perform in generating them? How are LLM-generated summaries different from those written by human experts? However, the lack of a comprehensive, high-quality scientific TLDR dataset hinders both the development and evaluation of LLMs' summarization ability. To address these, we propose a novel dataset, \dataname, containing a large sample of researcher-authored summaries from scientific papers, which leverages the common practice of including authors' comments alongside bibliography items. We then test popular open-weight LLMs for generating TLDRs based on abstracts. Our analysis reveals that, although some of them successfully produce humanoid summaries, LLMs generally exhibit a greater affinity for the original text's lexical choices and rhetorical structures, hence tend to be more extractive rather than abstractive in general, compared to humans. Our code and datasets are available at \url{https://github.com/netknowledge/LLM_summarization}~\cite{code}. 
\end{abstract}

\section{Introduction}

We, as researchers, are often called upon to condense our work into concise and informative summaries---whether it is a one-sentence teaser or a take-home message---across diverse contexts such as journal submissions and academic presentations. More broadly, high-quality extreme summaries, or TLDRs, facilitate the communication of scientific findings to diverse audiences, from domain experts to the lay public, which is essential for advancing research and promoting public engagement~\cite{jamieson2017oxford}.

\begin{figure}[t!]
    \centering
    \includegraphics[width=1\columnwidth]{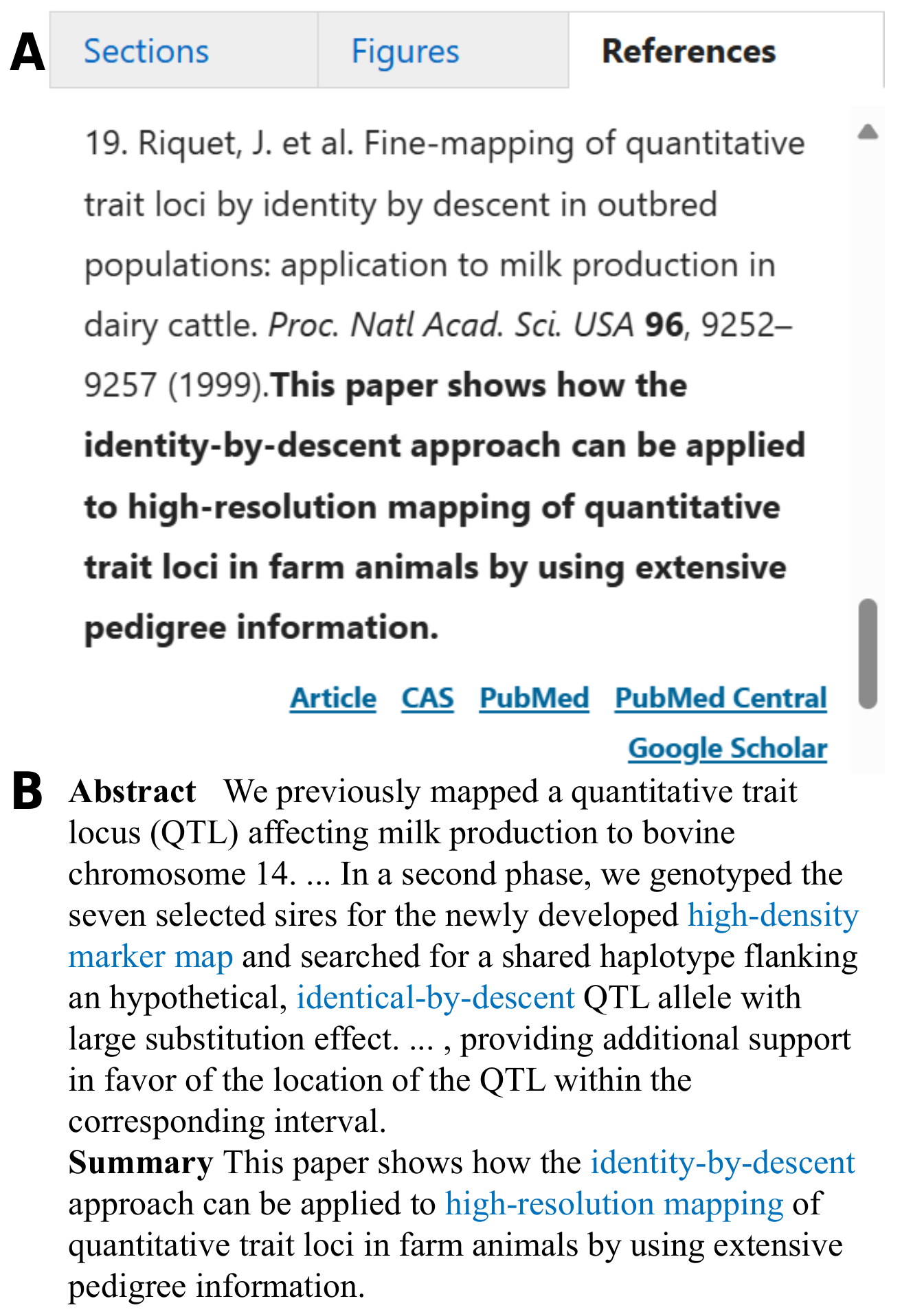}
    \caption{
    \textbf{(A)} An example annotated bibliography entry cited by \citet{andersson2001genetic}. Bold text at the end is the annotation from the author, which is used as the summary in this research. 
    \textbf{(B)} The corresponding abstract-summary pair to (A). The abstract is from \citet{riquet1999finemapping}. 
    }
    \label{fig:annotated_bibliography}
\end{figure}

Recently, Large Language Models (LLMs) have begun to transform this landscape, promising to automate the generation of these TLDRs~\cite{binz2025should}. However, the efficacy of LLMs in this domain is bottlenecked by the scarcity of high-quality scientific summary corpora. Predominant large-scale datasets often rely on extracting citation contexts from the ``Related Work" sections of citing papers. Such supervision signals are inherently fragmented, as they typically capture only specific facets of the cited work tailored to the citing author's narrative, failing to provide a holistic overview. Furthermore, their reliance on full-text extraction introduces significant availability barriers, as well as unnatural and imbalanced discipline distribution.

To overcome these obstacles, we introduce \dataname, a large-scale corpus derived from annotated bibliographies, which consists of citations accompanied by a few brief, descriptive sentences in which the author explicitly summarizes the referenced work, as shown in Fig.~\ref{fig:annotated_bibliography}A. This structural format provides a self-contained summary of the cited paper's core contributions, avoiding fragmentation and narrative bias. By systematically collecting these entries, we construct a dataset of over 35 thousand abstract-summary pairs, with most of the source abstracts and target summaries originating from highly cited papers published in top-tier biology or medical journals.

Using \dataname, we then conduct a multi-dimensional evaluation of state-of-the-art LLMs, uncovering a distinct behavioral divergence between LLM and human summarization. Our analysis reveals that while exhibiting a stronger extractive tendency, some of LLMs have reached human parity in identifying salient information, but they generally differ fundamentally in how present and rearrange it. 

Our contributions are summarized as follows:
\begin{enumerate}
    \item We propose a novel data collection paradigm based on annotated bibliographies and release \dataname, a dataset comprising high-quality researcher-authored TLDRs, aiming to establish a new benchmark for extreme scientific summarization.
    \item We provide a comprehensive evaluation framework that systematically quantifies the writing styles and tendencies of state-of-the-art LLMs.
\end{enumerate}

\section{Related Work}

Early research in automatic summarization primarily focused on newswire domains (e.g., CNN/DailyMail ~\cite{hermann2015teaching}). However, the exponential growth of scientific literature has created a pressing need for automated tools to mitigate information overload and enable rapid screening. These tools have proven highly effective in real-world applications, such as Semantic Scholar, which helps researchers speed up the process of targeting relevant papers by distilling complex research into a single, highly informative sentence. The scientific domain presents unique challenges due to its specialized terminology, lengthy documents, and the requirement for high-level domain expertise, which necessitates the construction of specialized scientific TLDR corpora distinct from traditional summarization datasets, for both the development and the evaluation of automatic summarization models.

\citet{volske-etal-2017-tl} pioneered this direction by mining user-generated summaries from Reddit, demonstrating the feasibility of learning extreme summarization from large-scale social media data. However, the informal and noisy nature of Reddit posts limits their direct applicability to rigorous scientific communication. Transposing this task to the scientific domain, \citet{cachola-etal-2020-tldr} introduced SciTLDR, a benchmark dataset derived from OpenReview. By combining author-written summaries with those curated by graduate students, SciTLDR established a high-quality gold standard. Nevertheless, its reliance on manual curation results in a relatively small scale and limited domain coverage.

To address the scalability bottleneck of manual annotation, recent efforts have turned to automated extraction methods. For example, \citet{mao2022citesum} proposed CiteSum, which extracts citation texts from the ``Related Work'' sections of papers as proxies for summaries. While this approach enables the construction of a large-scale dataset from the S2ORC corpus, it inherently relies on access to full-text articles, subjecting the dataset to copyright restrictions and potential disciplinary biases (80.04\% are computer science papers and only 2.13\% are medical papers). Moreover, citation texts often emphasize certain aspects pertinent to the citing work rather than providing a holistic summary of the cited paper. 

\begin{table*}[t!]
    \centering
    \caption{Statistics of existing scientific TLDR datasets, where ours is in bold. Average values are reported for words and sentences, whereas the median is used for citation counts.}
    \begin{threeparttable}
    \begin{tabular}{llrccc}
    \toprule
     \multirow{2}{*}{\textbf{Dataset}} & \multirow{2}{*}{\textbf{Domain}} & \multirow{2}{*}{\textbf{\# pairs}} & \textbf{\# words}& \textbf{\# sentences} & \textbf{\# citations}
     \\
     & & & \multicolumn{2}{c}{(source/target/ratio)} & (source/target)
     \\
     \midrule
     SciTLDR & OpenReview & 5,412 & 179.92/22.15/8.12 & 7.46/1.04/7.17 & - / -
     \\
     Multi-XScience & arXiv & 40,528 & 772.69/120.50/6.41 & 30.72/4.85/6.33 & 192/5
     \\
     CiteSum & S2ORC & 92,946 & 283.69/25.16/11.28 & 11.34/1.15/9.86 & 89/ - 
     \\
     \textbf{\dataname} & Springer & 35,626 & 191.58/25.29/7.58 & 7.37/1.14/6.46 & 308/435
     \\
     \bottomrule
    \end{tabular}
    \end{threeparttable}
    \label{tab:tldr-data}
\end{table*}

Beyond single-document extreme summarization, the research landscape has expanded into diverse input scenarios and objectives. \citet{lu2020multi} introduced Multi-XScience with emphasis on multi-document summarization, using a whole paragraph from a paper's ``Related Work'' section as the target text, and uses that paper's abstract, together with abstracts of all papers cited in that paragraph, as the source text. \citet{takeshita2022xscitldr} and \citet{atri2023fusing} extended the boundaries of the task to cross-lingual and multimodal settings, respectively, incorporating multiple language translations and audio-visual conference presentation data. A parallel line of research~\cite{goldsack-etal-2022-making} and \cite{guo2024retrieval} focuses on focused ``lay summarization'', aiming to make science accessible to the general public through expert-written simplifications and retrieval-augmented LLMs.

\section{Our Dataset}

We introduce a dataset comprising 35,626 paper abstracts paired with researcher-authored summaries. This resource aims to enrich academic-targeted training corpora and facilitate evaluations of LLMs' summarization ability by establishing a human baseline. 

To construct the dataset, we systematically crawl web pages for over 1.2 million academic papers from Springer's websites, with an example shown in Fig.~\ref{fig:annotated_bibliography}A. Then we parse retrieved HTML files to extract bold annotations and the DOI of corresponding citable items. For most pages, a Google Scholar link is provided for each reference, from which we obtain the embedded DOI; when the link is absent, we use CrossRef's API to query the DOI based on the bibliography text. After collecting DOI-summary pairs, we use DOIs to match the abstracts of corresponding academic papers (as shown in Fig.~\ref{fig:annotated_bibliography}B), resulting in a dataset consisting of 35,626 abstract-summary pairs involving 29,356 papers. Some samples from our dataset are available in \S~\ref{sec:llm_sum_sample}. 

Table~\ref{tab:tldr-data} provides a quantitative comparison of several existing scientific TLDR datasets. Our \dataname dataset distinguishes itself from them in several qualitative dimensions. First, our use of annotated bibliographies---commonly found in highly cited review papers authored by experienced researchers---ensures a more authoritative and standalone overview of the referenced work. Moreover, by relying solely on universally accessible abstracts and bibliography metadata, our approach circumvents the copyright restrictions and parsing noise associated with full-text extraction, thereby enhancing reproducibility and disciplinary coverage, as shown in \S~\ref{subsec:discipline}. 

\section{Generating Summary using LLMs}

To measure LLMs' summarization ability and to examine how LLM-generated summaries differ from those written by domain experts, we set up a local Ollama
server which hosts 5 open-weights LLMs, namely \texttt{gemma3:27b}~\cite{team2025gemma}, \texttt{deepseek-r1:70b}~\cite{deepseek-ai2025deepseekr1}, \texttt{gpt-oss:12b}~\cite{openai2025gptoss120b}, \texttt{llama4:16x17b}~\cite{llama}, and \texttt{qwen3:235b}~\cite{yang2025qwen3}. We employ a few-shot learning framework that guides LLMs through summarization (see \S~\ref{sec:llm_gen_sum_prompt}). Specifically, 5 randomly selected and then fixed abstract-summary pairs are provided for LLMs, along with the word count of each summary, as the training set. The other 35,621 pairs serve as the test set. Then, we feed LLMs with abstracts and word counts of human-written summaries as length reference and ask them to generate summarizations recursively on the test set. Samples in both the training and test sets are identical for all LLMs. We set the context window size to 4096 and the temperature to 1 for all LLMs. We disabled DeepSeek-R1, gpt-oss, and Qwen3's thinking mode due to the complexity of the processing speed required for this task. Every abstract in the test set was summarized individually, and we got $178,105$ ($35,621\times5$) LLM-generated summaries in total. Some samples of them can be checked in \S~\ref{sec:llm_sum_sample}.

\section{Evaluation}

We present multi-dimensional comparisons between the extreme scientific summaries in our dataset, their respective abstracts, and LLM-generated summaries. These analyses include summary statistics (\S~\ref{subsec:summary}), readability (\S~\ref{subsec:readability}), abstractiveness (\S~\ref{subsec:abstractiveness}), rhetorical structure (\S~\ref{subsec:rhetorical}), and reference-based evaluation (\S~\ref{subsec:ref_based}), which collectively aim to understand the differences between different types of summaries. 

\begin{table*}[t!]
\centering
\caption{Statistics for summaries by LLMs and human researchers, compared with original abstracts. Reported are mean~$\pm$~standard deviation.}
\label{tab:sum_stat}
\resizebox{\linewidth}{!}{%
\begin{tabular}{l|ccc|c|ccc}
\toprule
\multirow{2}{*}{\textbf{Model}}&\multicolumn{3}{c|}{\textbf{Length}}&\multirow{2}{*}{\textbf{\makecell{NEs per \\100 words}}}&\multicolumn{3}{c}{\textbf{Training set copy rate (\%)}}\\
& words& sentences & words/sent.&&$=0\%$&$>10\%$&Max\\
\midrule
Gemma 3&$29.06\pm11.03$&$1.03\pm2.31$&$28.31\pm9.59$&$4.49\pm4.64$&91.25&0.28&27.78\\
DeepSeek-R1&$60.46\pm158.34$&$2.31\pm7.34$&$26.19\pm8.30$&$4.13\pm3.71$&90.83&0.19&100.00\\
Gpt-oss&$62.35\pm22.84$&$1.64\pm0.93$&$37.94\pm15.51$&$\mathbf{6.30}\pm4.60$&\textbf{96.16}&\textbf{0.01}&\textbf{11.48}\\
Llama 4&$43.11\pm764.99$&$1.30\pm3.32$&$33.24\pm628.57$&$3.46\pm4.28$&86.63&1.05&46.15\\
Qwen3&$153.29\pm1799.10$&$6.22\pm91.82$&$24.64\pm140.67$&$4.90\pm4.38$&94.88&0.03&95.24\\
\midrule
Human &$25.29\pm11.43$&$1.14\pm.41$&$22.22\pm8.83$&$5.56\pm5.93$ &-&-&-\\
\midrule
Abstract&$191.58\pm82.52$&$7.37\pm3.18$&$26.01\pm11.71$&$5.11\pm3.83$&-&-&-\\
\bottomrule
\end{tabular}
}
\end{table*}

\subsection{Descriptive Statistics} \label{subsec:summary}

Table~\ref{tab:sum_stat} presents some summary statistics of LLM-generated summaries, human-written summaries, as well as the original abstracts for reference. All LLMs, except Gemma 3, produce summaries whose lengths significantly exceed the intended length, despite providing a reference word count during training. Furthermore, LLMs tend to generate repetitive or verbose content to varying degrees (see \S~\ref{sec:nonsense}), reflected in the extremely large standard deviations of Qwen3 and Llama 4's summary lengths. 

Identifying named entities using the pretrained \verb|en_core_web_sm| model from spaCy~\cite{Honnibal_spaCy_Industrial-strength_Natural_2020}, we find that gpt-oss exhibits the most distinguishable density of information content in its generated summary, reflecting a higher specificity. The other LLM-generated summaries show lower entity density compared to human-written ones in general. 

We evaluate the extent LLM-generated summaries copy phrases from the training set examples, assessing possible overfitting or memorization. Specifically, we define the training set copy rate of a summary as the percentage of its 3-grams that overlap with those in the 5 training samples. Table~\ref{tab:sum_stat} reports 3 statistics: the proportion of summaries with a $0\%$ copy rate, the proportion with a copy rate exceeding $10\%$, and the maximum copy rate observed across all summaries. The results show that all LLMs exhibit a varying extent of copying from the training set.

\subsection{Readability} \label{subsec:readability}

Ideally, an effective TLDR should lower the cognitive barrier for readers, transforming complex technical content into plain language that is easier to digest. To assess the linguistic complexity and accessibility of the generated summaries, we employ 7 established readability metrics: Flesch-Kincaid Grade (FKG), Gunning Fog (GF), SMOG Index (SI), Automated Readability Index (ARI), Coleman-Liau Index (CLI), Linsear Write Formula (LWF), and Dale-Chall Readability Score (DCRS). These metrics estimate the US grade level required to comprehend a text by analyzing features such as sentence length, syllable count, and the proportion of complex or polysyllabic words. For all metrics, lower scores indicate higher readability and greater accessibility to a broader audience.

\begin{table}[t!]
    \centering
    \caption{Comparison of mean readability scores across different models and human baselines. For all metrics, lower values denote better readability. }
    \label{tab:readability_scores}
    \resizebox{\columnwidth}{!}{%
    \begin{tabular}{lccccccc}
    \toprule
    \textbf{Model} & \textbf{FKG} $\downarrow$ & \textbf{GF} $\downarrow$ & \textbf{SI} $\downarrow$ & \textbf{ARI} $\downarrow$ & \textbf{CLI} $\downarrow$ & \textbf{LWF} $\downarrow$ & \textbf{DCRS} $\downarrow$ \\
    \midrule
    Gemma 3 & 19.20 & 23.46 & 19.52 & 21.68 & 19.55 & 21.12 & 14.48 \\
    DeepSeek-R1 & 19.69 & 23.85 & 19.76 & 22.47 & 20.80 & 21.02 & 14.94 \\
    Gpt-oss & 23.79 & 27.82 & 22.61 & 28.45 & 21.55 & 29.90 & 15.77 \\
    Llama 4 & 18.13 & 21.80 & 18.25 & 20.22 & 17.58 & 19.86 & 13.54 \\
    Qwen3 & 22.94 & 27.27 & 22.05 & 26.52 & 23.15 & 25.70 & 15.94 \\
    \midrule
    Human & \textbf{15.25} & \textbf{18.79} & \textbf{16.13} & \textbf{16.58} & \textbf{16.08} & \textbf{15.70} & \textbf{13.05} \\
    \midrule
    Abstract & 16.17 & 19.75 & 17.18 & 17.91 & 16.61 & 16.38 & 13.65 \\
    \bottomrule
    \end{tabular}}
\end{table}

Table~\ref{tab:readability_scores} reveals a distinct divide regarding readability between LLM-generated summaries and human-written ones. Surprisingly, we find that all LLM-generated summaries exhibit significantly higher scores across all metrics compared to not only human-written summaries but also the original abstracts. Among LLMs, Llama 4 proves to be the most readable model, despite its tendency to generate longer summaries.  On the other hand, summaries from human researchers are consistently more readable than the corresponding abstracts, despite the small scale of the difference. We perform two-sided Mann-Whitney U tests and find that the differences between all summaries and the original abstract are statistically significant (all $p-$values are smaller than 0.001).

We hypothesize that this discrepancy stems from the tendency of LLMs to excessively rely on high-density domain-specific jargon. In contrast, human researchers likely exercise greater audience awareness during the summarization process; they may intentionally adopt plain language to ensure their summaries remain accessible to reviewers or readers from outside the specific domain \cite{devaraj2021paragraphlevel}. We attribute this to its lower density of named entities (as shown in Table~\ref{tab:sum_stat}), a factor that inversely correlates with the poor readability of gpt-oss.

\subsection{Abstractiveness} \label{subsec:abstractiveness}

While extractive summarization primarily selects and rearranges existing sentences, an effective TLDR often requires abstractive capabilities to paraphrase complex concepts and synthesize information concisely. To evaluate this, following previous studies \cite{sharma2019bigpatent, see2017get}, we calculate the proportion of $n$-grams in a summary that do not appear in the original source abstract (i.e., novel $n$-grams) as a measure of abstractiveness. A higher value indicates that the summary employs more original phrasing rather than simply copying chunks of text from the abstract, i.e., to be more abstractive rather than extractive, reflecting a higher level of lexical novelty and creativity exhibited by the writer. 

\begin{table}[t!]
    \centering
    \caption{Proportion of novel $n$-grams in LLM-generated summaries compared to human-written ones. Values are reported as mean~$\pm$~standard deviation. Higher values indicate more original vocabulary usage.}
    \label{tab:novel_ngrams}
    \resizebox{\columnwidth}{!}{%
    \begin{tabular}{lccc}
    \toprule
    \multirow{2}{*}{\textbf{Model}} & \multicolumn{3}{c}{\textbf{\% novel $n$-grams}} \\
    \cmidrule(lr){2-4}
     & $n=1$ & $n=2$ & $n=3$ \\
    \midrule
    Gemma 3 & $35.44\pm11.35$ & $77.12\pm12.49$ & $90.20\pm9.90$\\
    DeepSeek-R1 & $33.48\pm9.59$ & $73.81\pm10.13$ & $87.90\pm8.20$\\
    Gpt-oss & $39.22\pm9.36$ & $80.94\pm7.95$ & $92.70\pm5.64$\\
    Llama 4 & $39.85\pm13.48$ & $81.63\pm13.23$ & $92.67\pm10.21$\\
    Qwen3 & $37.66\pm10.39$ & $80.74\pm9.33$ & $92.91\pm6.78$\\
    \midrule
    \textbf{Human} & $\mathbf{46.13}\pm14.23$ & $\mathbf{86.33}\pm13.16$ & $\mathbf{95.09}\pm9.86$\\
    \bottomrule
    \end{tabular}
    }
\end{table} 

Table~\ref{tab:novel_ngrams} illustrates that human writers consistently exhibit a greater tendency to introduce new vocabulary not present in the source text compared to all LLMs, suggesting a significantly superior abstractiveness. Among the models, Llama 4 and gpt-oss demonstrate the highest level of abstractiveness. By contrast, smaller or more constrained models like Gemma 3 and DeepSeek-R1 show lower novelty scores, indicating a stronger reliance on extractive copying from the source abstracts. These results suggest that while advanced LLMs are becoming increasingly capable of paraphrasing, human researchers still maintain a distinct edge in synthesizing information with original language.

\subsection{Rhetorical Structure} \label{subsec:rhetorical}

Understanding the rhetorical structure of summaries---how information is organized across functional roles such as background, objective, methods, results, and conclusions---provides insight into how writers prioritize and sequence content. This is particularly relevant for TLDR, where space constraints force selective emphasis on certain rhetorical elements.

To analyze rhetorical structure, we employ a DeepSeek-R1-based classifier to assign each sentence in the original abstracts to one of 5 rhetorical roles: \textsc{Background}, \textsc{Objective}, \textsc{Methods}, \textsc{Results}, and \textsc{Conclusions} (see \S~\ref{sec:appendix_rhetorical} for details). For each word $w$ appearing at a normalized position $p$ in a summary, we compute the probability of its rhetorical role based on word occurrence statistics from the corresponding abstract. Specifically, let $c_{w,r}$ denote the count of word $w$ in sentences labeled with role $r$, and $N_r$ the total word count in role $r$. We define the normalized probability as: 
\begin{equation}
    Pr(r \mid w) = \frac{c_{w,r} / N_r}{\sum_{r'} c_{w,r'} / N_{r'}} \, .
\end{equation}
For words in the summary that do not appear in the corresponding abstract, we assign them to a special \textsc{Novel} category with $Pr (\textsc{Novel} \mid w) = 1$, indicating content introduced beyond the source abstract. We then aggregate these probabilities across all summaries by binning word positions into 20 equal intervals.

\begin{figure}[t!]
    \centering
    \includegraphics[width=1\columnwidth]{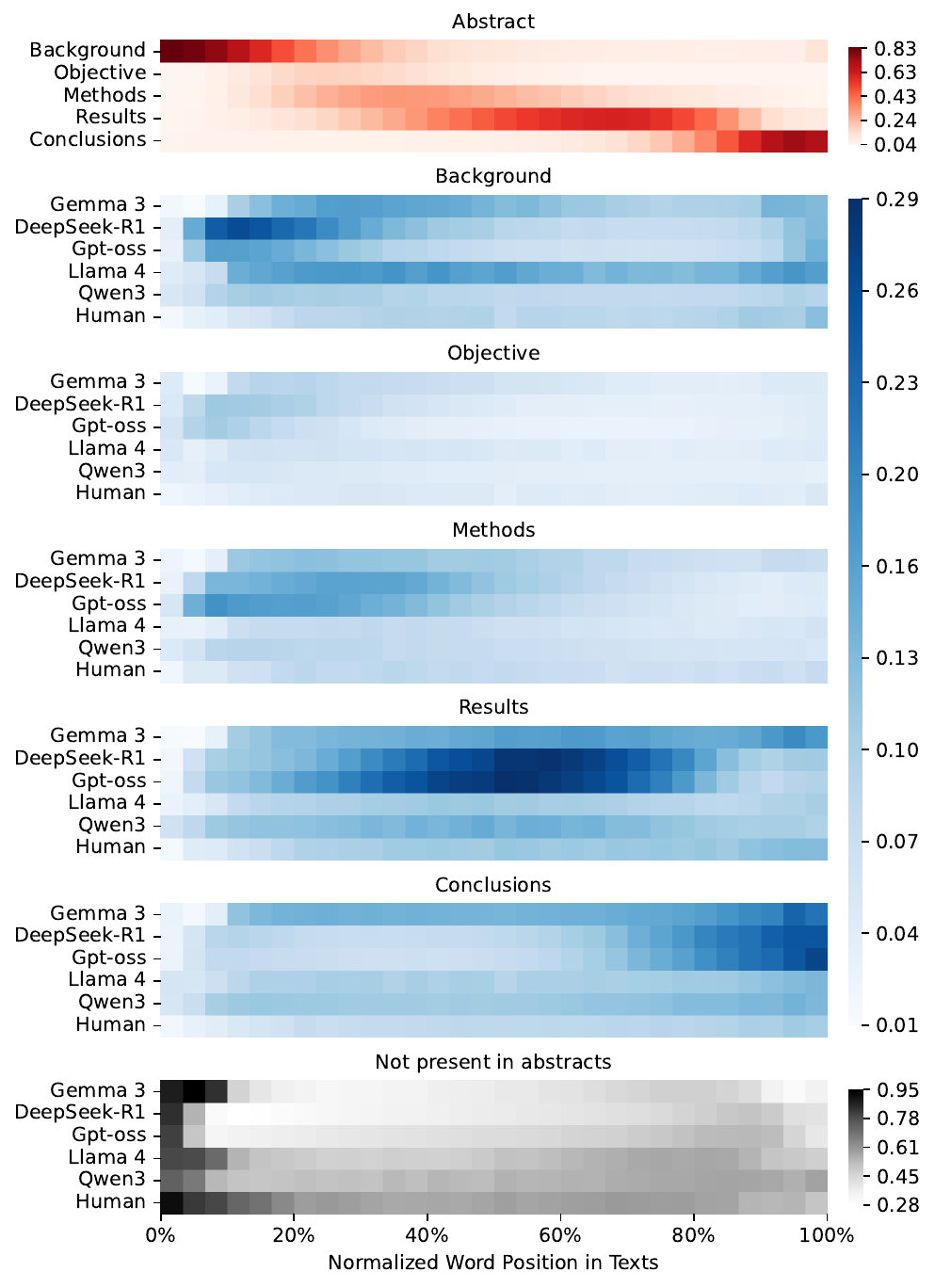}
    \caption{Word rhetorical role distribution across different types of summaries, and original abstracts as the reference. The $x$-axis represents normalized word position in texts, and color intensity indicates the average probability of belonging to each rhetorical role, based on word co-occurrence statistics against the original abstract.}
    \label{fig:rhetorical}
\end{figure}

Figure~\ref{fig:rhetorical} visualizes the resulting rhetorical role distributions, where darker colors indicate higher probabilities that words at a given position originate from a specific rhetorical section of the abstract. For example, if the \textsc{Background} row shows dark blue at the beginning, it suggests that words in the opening of summaries are predominantly drawn from background-related sentences in the source abstract.

Several patterns emerge from this analysis. First, all text sources---LLM-generated summaries, human summaries, and original abstracts---exhibit a similar overall progression from \textsc{Background} toward \textsc{Conclusions}, reflecting the conventional structure of scientific discourse, while DeepSeek-R1 and gpt-oss display a closer mirroring of the rhetorical structure, when compared to the original abstracts. Second, across all summary types, words are predominantly drawn from the \textsc{Results} sections of the original abstracts, with \textsc{Objective} contributing the least, indicating a shared preference for reporting findings over restating research goals. Third, we observe that all types of summaries introduce a substantial number of novel words---words not present in the original abstracts---at the beginning positions. Upon inspection, these are predominantly meta-discourse markers such as ``study,'' ``demonstrates,'' ``identifies,'' and ``shows'' (see Figure~\ref{fig:top10_novel_words}), which serve to frame the research in line with research norms. Fourth, compared to the relatively concentrated distributions of \textsc{Background} and \textsc{Conclusions} in original abstracts, summaries tend to spread these roles more broadly across positions, particularly for Gemma 3 and Llama 4, and human, indicating a potentially more flexible functionality structure.

\subsection{Reference-based Evaluation} \label{subsec:ref_based}

We benchmark LLMs against human experts to assess summarization quality. Following the widely used evaluation methodology, we compare generated summaries to reference ones using lexical overlap-based measures, including BLEU~\cite{papineni2002bleu}, $F_1$~ROUGE~\cite{lin2004rouge}, and METEOR~\cite{banerjee2005meteor}, as well as semantic similarity-based measures that capture deeper semantic correspondence, including BERTScore~\cite{zhang2019bertscore} and MoverScore~\cite{zhao2019moverscore}. 

Using researcher-written summaries as references, we find that Gemma 3 and DeepSeek-R1 perform best on semantic similarity (MoverScore) and lexical overlap (METEOR), respectively (upper half of Table~\ref{tab:ref-based}). These metrics-based results are best interpreted when positioning human-authored summaries as a counterpart for comparison. Specifically, we leverage the fact that papers in our dataset can receive multiple summaries (typically 2-10; see Figure~\ref{fig:paper_annotation_counts}). To establish a human performance baseline, we conduct a within-paper, cross-referenced evaluation of human-authored summaries with 10 iterations. In each iteration, we randomly select one summary as the reference and calculate metrics for the remaining summaries. The reported values represent averages across these iterations. The results indicate that human authors achieve the highest scores on BLEU, ROUGE, and BERTScore metrics (the sixth row in Table~\ref{tab:ref-based}), suggesting that human-written summaries of the same paper exhibit stronger $n$-gram exact matches and semantic consistency than those generated by LLMs. However, the high scores on METEOR and MoverScore metrics indicate that models like Gemma 3 and DeepSeek-R1 are highly effective at identifying the same salient information as human experts, by successfully capturing the gist and core technical entities that humans prioritize.

\begin{table*}[t!]
\centering
\caption{Evaluation of summarization performance across LLMs and human authors using two references: researcher-written summaries and original paper abstracts.}
\label{tab:ref-based}
\begin{tabular}{l|ccc|cc}
\toprule
\multirow{2}{*}{\textbf{Model}}&\multicolumn{3}{c|}{\textbf{Overlap-based}}&\multicolumn{2}{c}{\textbf{Similarity-based}}\\
&BLEU&ROUGE(1/2/L)&METEOR&BERTScore&MoverScore\\
\midrule
\multicolumn{6}{c}{\textbf{Human-referenced performance}} \\
\midrule
Gemma 3&.0290&.2634/.0584/.2021&.2216&.8817&\textbf{.5660}\\
DeepSeek-R1&.0179&.2375/.0514/.1703&\textbf{.2486}&.8715&.5562\\ 
Gpt-oss&.0131&.2233/.0443/.1549&.2255&.8592&.5489\\
Llama 4&.0188&.2458/.0508/.1865&.2090&.8771&.5593\\
Qwen3&.0046&.2272/.0465/.1665&.2129&.8699&.5578\\
Human&\textbf{.0450} & \textbf{.2734/.0675/.2163} & .2155 & \textbf{.8824} & .5642\\
\midrule
\multicolumn{6}{c}{\textbf{Abstract-referenced performance}} \\
\midrule
Gemma 3&.0006&.1931/.0752/.1381&.1004&.8637&.5524\\
DeepSeek-R1&.0226&.3279/\textbf{.1434}/.2306&\textbf{.1974}&\textbf{.8838}&\textbf{.5836}\\ 
Gpt-oss&.0167&\textbf{.3436}/.1323/\textbf{.2347}&.1788&.8751&.5817\\ 
Llama 4&.0035&.1878/.0616/.1263&.0942&.8565&.5438\\
Qwen3&\textbf{.0307}&.2495/.0911/.1659&.1345&.8694&.5681\\
Human&.0001&.1409/.0382/.1005&.0638&.8410&.5302\\
\bottomrule
\end{tabular}
\end{table*}

A contrasting pattern emerges when evaluating against the original abstracts. Human-authored summaries show both the lowest overlap and similarity scores (lower half of Table~\ref{tab:ref-based}), suggesting that human annotators tend to substantially reformulate content and introduce novel phrasing. This might also stem from human authors drawing on full texts rather than abstracts alone for summarization. Among LLMs, Qwen3 and gpt-oss attain the highest BLEU and ROUGE scores, respectively, indicating their different inclinations in word choice and the trade-off between precision and recall. While DeepSeek-R1 achieves both the highest METEOR and similarity scores, surpassing gpt-oss and Qwen3, indicating that DeepSeek-R1 maintains the closest lexical and semantic fidelity to the source abstract, even with fewer parameters. 

These contrasting results presented in Table~\ref{tab:ref-based} highlight an important, well-recognized tension in summarization: extractive versus abstractive~\cite{durmus-etal-2020-feqa}. Human researchers appear to prioritize the latter, whereas LLMs tend to emphasize the former by exhibiting stronger adherence to the source text, which aligns with Table~\ref{tab:novel_ngrams}.

\section{Discussion}

Our comprehensive evaluations reveals distinct behavioral profiles across LLMs, suggesting the need to align model selection for scientific summarization with  user priorities. For approximating human behaviors, Gemma 3 emerges as a balanced candidate, uniquely adhering to length constraints while achieving high semantic similarity to expert-written references. If precise, high-fidelity condensation is the goal, DeepSeek-R1 and gpt-oss maintain strict faithfulness to the original vocabulary and rhetorical structure, though via distinct mechanisms: DeepSeek-R1 adopts a conservative, extractive approach. Its superior METEOR score evidences a precise alignment with researchers in terminology preferences and content selection, yet its low lexical novelty suggests a minor level of paraphrasing. In contrast, gpt-oss functions as a high-fidelity rewriter; it achieves the highest content recall (ROUGE) alongside substantial lexical novelty, effectively rephrasing the source text without sacrificing factual detail. Llama 4 presents a different trade-off, offering the highest readability scores suitable for non-expert audiences, as well as the highest lexical novelty, yet it suffers from significant hallucinations in length control. Finally, Qwen3 demonstrates strong lexical precision (high BLEU) yet its practical utility is compromised by significant instability and a tendency toward verbosity.

Regarding model size, an interesting observation is that larger models do not necessarily outperform smaller ones in summarization. On the contrary, lighter models such as Gemma 3, DeepSeek-R1, and gpt-oss demonstrate stronger capabilities and greater similarity to human summaries. Notably, \verb|deepseek-r1:70b|---a distilled model of \verb|llama3:70b| fine-tuned on reasoning data generated by \verb|deepseek-r1:671b|---even surpasses the base model's next generation \verb|llama4:16x17b|, which has more parameters. This result underscores the value of model distillation and data-centric fine-tuning~\cite{deepseek-ai2025deepseekr1}, prompting us to rethink the potential of parameter-efficient models and move beyond the conventional scaling law that favors increasing model size.

Beyond model performance, this study validates a novel and scalable pipeline for constructing high-quality scientific summarization datasets. By leveraging annotated bibliographies, where experts manually curate and summarize the literature, we circumvent the high demand for domain knowledge in large-scale annotation. This methodology is inherently generalizable since it can be readily adapted to any academic archive or indexing service that maintains bibliographic metadata. Moreover, by relying solely on abstracts and citation metadata, which are typically open-access or fall under fair use, our approach mitigates copyright concerns and lowers barriers for future research in NLP.

\section{Conclusion}

In this work, we present \dataname, a scalable summary corpus that provides a comprehensive disciplinary coverage aligned with the actual publication landscape, offering a high-quality, academic context tailored alternative to existing TLDR datasets. Using \dataname, we conduct a multi-dimensional evaluation of current LLMs and highlight that LLMs remain more extractive compared to human authors. Among the models, DeepSeek-R1 archives the highest fidelity to source texts, gpt-oss balances faithfulness with lexical novelty, and Gemma 3 uniquely prioritizes semantic alignment, mimicking the synthesis style of human researchers without relying on extensive lexical overlap.

\section*{Limitations}

We acknowledge several limitations of our work. First, although the dataset is high quality, its source---annotated bibliographies---is not a perfect proxy for TLDR. Annotation styles and intentions can vary, reflecting personal perspectives or critiques rather than purely objective summaries. Nevertheless, the dataset's large scale enables a unique perspective into how LLMs perform extreme summarization relative to human authors. 

Second, while LLMs were restricted to using only paper abstracts as input, human annotators may likely draw on abstracts and full texts, which could partly explain the observed low abstract-referenced similarity. Future work could explore whether providing LLMs with full-text would lead to further improvements in summary quality. 

Third, although we explicitly constrained LLM output length in the prompt, most generated summaries were significantly longer than those by human researchers. This calls for dedicated experiments to rigorously decouple the relationship between summary quality and length, despite evaluation metrics (e.g., ROUGE, METEOR) claiming robustness to length variation. 

Fourth, our dataset is not fully representative of the entire scientific literature in terms of citations, as all reported performance is based on highly cited works. Therefore, evaluating LLMs summarization on the general population of papers remains as an important direction for future research. 

Finally, we evaluated only open-source models and selected their smaller variants, due to computational and cost constraints. Consequently, our study excludes proprietary models or the largest architectures (e.g., \verb|DeepSeek-r1:671b| and \verb|llama4:128x17b|), limiting a comprehensive performance comparison across the full spectrum of current LLM capabilities.

\section*{Ethical considerations}

Our experiments reveal significant instability in LLM outputs, posing challenges for their practical adoption in scientific workflows, as evidenced by the extreme standard deviations in summary length and the non-trivial rates of training set memorization. When coupled with our finding that model-generated summaries can be less readable than the source abstracts, these inconsistencies suggest that LLMs cannot yet be fully automated as unsupervised research assistants. Instead, they require rigorous human oversight and post-hoc verification to ensure the integrity and accessibility of scientific communication.

\section*{Acknowledgments}

\bibliography{not_too_long_do_read}

@inproceedings{hermann2015teaching,
  title = {Teaching {{Machines}} to {{Read}} and {{Comprehend}}},
  booktitle = {Advances in {{Neural Information Processing Systems}}},
  author = {Hermann, Karl Moritz and Kocisky, Tomas and Grefenstette, Edward and Espeholt, Lasse and Kay, Will and Suleyman, Mustafa and Blunsom, Phil},
  year = 2015,
  volume = {28},
  publisher = {Curran Associates, Inc.}
}

@inproceedings{atri2023fusing,
author = {Atri, Yash Kumar and Goyal, Vikram and Chakraborty, Tanmoy},
title = {Fusing Multimodal Signals on Hyper-complex Space for Extreme Abstractive Text Summarization (TL;DR) of Scientific Contents},
year = {2023},
isbn = {9798400701030},
publisher = {Association for Computing Machinery},
address = {New York, NY, USA},
url = {https://doi.org/10.1145/3580305.3599830},
doi = {10.1145/3580305.3599830},
booktitle = {Proceedings of the 29th ACM SIGKDD Conference on Knowledge Discovery and Data Mining},
pages = {3724–3736},
numpages = {13},
location = {Long Beach, CA, USA},
series = {KDD '23}
}

@inproceedings{takeshita2022xscitldr,
author = {Takeshita, Sotaro and Green, Tommaso and Friedrich, Niklas and Eckert, Kai and Ponzetto, Simone Paolo},
title = {X-SCITLDR: cross-lingual extreme summarization of scholarly documents},
year = {2022},
isbn = {9781450393454},
publisher = {Association for Computing Machinery},
address = {New York, NY, USA},
url = {https://doi.org/10.1145/3529372.3530938},
doi = {10.1145/3529372.3530938},
booktitle = {Proceedings of the 22nd ACM/IEEE Joint Conference on Digital Libraries},
articleno = {4},
numpages = {12},
location = {Cologne, Germany},
series = {JCDL '22}
}

@inproceedings{mao2022citesum,
    title = "{C}ite{S}um: Citation Text-guided Scientific Extreme Summarization and Domain Adaptation with Limited Supervision",
    author = "Mao, Yuning  and
      Zhong, Ming  and
      Han, Jiawei",
    editor = "Goldberg, Yoav  and
      Kozareva, Zornitsa  and
      Zhang, Yue",
    booktitle = "Proceedings of the 2022 Conference on Empirical Methods in Natural Language Processing",
    month = dec,
    year = "2022",
    address = "Abu Dhabi, United Arab Emirates",
    publisher = "Association for Computational Linguistics",
    url = "https://aclanthology.org/2022.emnlp-main.750/",
    doi = "10.18653/v1/2022.emnlp-main.750",
    pages = "10922--10935"
}

@inproceedings{lu2020multi,
    title = "Multi-{XS}cience: A Large-scale Dataset for Extreme Multi-document Summarization of Scientific Articles",
    author = "Lu, Yao  and
      Dong, Yue  and
      Charlin, Laurent",
    editor = "Webber, Bonnie  and
      Cohn, Trevor  and
      He, Yulan  and
      Liu, Yang",
    booktitle = "Proceedings of the 2020 Conference on Empirical Methods in Natural Language Processing (EMNLP)",
    month = nov,
    year = "2020",
    address = "Online",
    publisher = "Association for Computational Linguistics",
    url = "https://aclanthology.org/2020.emnlp-main.648/",
    doi = "10.18653/v1/2020.emnlp-main.648",
    pages = "8068--8074"
}

@misc{code,
  title={LLM-generated extreme scientific summaries are preferred by machine learning models.},
  author={Zhuoqi Lyu and Qing Ke}, 
  year={2025},
  note={\url{https://github.com/netknowledge/LLM_summarization}. Deposited 12 August 2025.}
}

@book{jamieson2017oxford,
  title={The Oxford handbook of the science of science communication},
  author={Jamieson, Kathleen Hall and Kahan, Dan and Scheufele, Dietram A},
  year={2017},
  publisher={Oxford University Press},
  doi = {10.1093/oxfordhb/9780190497620.001.0001}
}

@article{binz2025should,
  title={How should the advancement of large language models affect the practice of science?},
  author={Binz, Marcel and Alaniz, Stephan and Roskies, Adina and Aczel, Balazs and Bergstrom, Carl T and Allen, Colin and Schad, Daniel and Wulff, Dirk and West, Jevin D and Zhang, Qiong and Shiffrin, Richard M. and Gershman, Samuel J. and Popov, Vencislav and Bender, Emily M. and Marelli, Marco and Botvinick, Matthew M. and Akata, Zeynep and Schulz, Eric},
  journal={Proceedings of the National Academy of Sciences},
  volume={122},
  number={5},
  pages={e2401227121},
  year={2025},
  doi={10.1073/pnas.2401227121}
}

@article{andersson2001genetic,
  title = {Genetic Dissection of Phenotypic Diversity in Farm Animals},
  author = {Andersson, Leif},
  year = 2001,
  journal = {Nature Reviews Genetics},
  volume = {2},
  number = {2},
  pages = {130--138},
  publisher = {Nature Publishing Group},
  doi = {10.1038/35052563}
}

@article{riquet1999finemapping,
  title = {Fine-Mapping of Quantitative Trait Loci by Identity by Descent in Outbred Populations: {{Application}} to Milk Production in Dairy Cattle},
  shorttitle = {Fine-Mapping of Quantitative Trait Loci by Identity by Descent in Outbred Populations},
  author = {Riquet, Juliette and Coppieters, Wouter and Cambisano, Nadine and Arranz, Juan-Jos{\'e} and Berzi, Paulette and Davis, Scott K. and Grisart, Bernard and Farnir, Fr{\'e}d{\'e}rick and Karim, Latifa and Mni, Myriam and Simon, Patricia and Taylor, Jeremy F. and Vanmanshoven, Pascal and Wagenaar, Danny and Womack, James E. and Georges, Michel},
  year = 1999,
  journal = {Proceedings of the National Academy of Sciences},
  volume = {96},
  number = {16},
  pages = {9252--9257},
  publisher = {Proceedings of the National Academy of Sciences},
  doi = {10.1073/pnas.96.16.9252}
}

@article{guo2024retrieval,
  title={Retrieval augmentation of large language models for lay language generation},
  author={Guo, Yue and Qiu, Wei and Leroy, Gondy and Wang, Sheng and Cohen, Trevor},
  journal={Journal of Biomedical Informatics},
  volume={149},
  pages={104580},
  year={2024},
  doi={10.1016/j.jbi.2023.104580}
}

@inproceedings{durmus-etal-2020-feqa,
    title = "{FEQA}: A Question Answering Evaluation Framework for Faithfulness Assessment in Abstractive Summarization",
    author = "Durmus, Esin  and
      He, He  and
      Diab, Mona",
    editor = "Jurafsky, Dan  and
      Chai, Joyce  and
      Schluter, Natalie  and
      Tetreault, Joel",
    booktitle = "Proceedings of the 58th Annual Meeting of the Association for Computational Linguistics",
    month = jul,
    year = "2020",
    address = "Online",
    publisher = "Association for Computational Linguistics",
    url = "https://aclanthology.org/2020.acl-main.454/",
    doi = "10.18653/v1/2020.acl-main.454",
    pages = "5055--5070",
}

@inproceedings{cachola-etal-2020-tldr,
    title = "{TLDR}: Extreme Summarization of Scientific Documents",
    author = "Cachola, Isabel  and
      Lo, Kyle  and
      Cohan, Arman  and
      Weld, Daniel",
    editor = "Cohn, Trevor  and
      He, Yulan  and
      Liu, Yang",
    booktitle = "Findings of the Association for Computational Linguistics: EMNLP 2020",
    month = nov,
    year = "2020",
    address = "Online",
    publisher = "Association for Computational Linguistics",
    url = "https://aclanthology.org/2020.findings-emnlp.428/",
    doi = "10.18653/v1/2020.findings-emnlp.428",
    pages = "4766--4777",
}

@inproceedings{volske-etal-2017-tl,
    title = "{TL};{DR}: Mining {R}eddit to Learn Automatic Summarization",
    author = {V{\"o}lske, Michael  and
      Potthast, Martin  and
      Syed, Shahbaz  and
      Stein, Benno},
    editor = "Wang, Lu  and
      Cheung, Jackie Chi Kit  and
      Carenini, Giuseppe  and
      Liu, Fei",
    booktitle = "Proceedings of the Workshop on New Frontiers in Summarization",
    month = sep,
    year = "2017",
    address = "Copenhagen, Denmark",
    publisher = "Association for Computational Linguistics",
    url = "https://aclanthology.org/W17-4508/",
    doi = "10.18653/v1/W17-4508",
    pages = "59--63"
}

@inproceedings{goldsack-etal-2022-making,
    title = "Making Science Simple: Corpora for the Lay Summarisation of Scientific Literature",
    author = "Goldsack, Tomas  and
      Zhang, Zhihao  and
      Lin, Chenghua  and
      Scarton, Carolina",
    editor = "Goldberg, Yoav  and
      Kozareva, Zornitsa  and
      Zhang, Yue",
    booktitle = "Proceedings of the 2022 Conference on Empirical Methods in Natural Language Processing",
    month = dec,
    year = "2022",
    address = "Abu Dhabi, United Arab Emirates",
    publisher = "Association for Computational Linguistics",
    url = "https://aclanthology.org/2022.emnlp-main.724/",
    doi = "10.18653/v1/2022.emnlp-main.724",
    pages = "10589--10604",
}

@article{team2025gemma,
  title={Gemma 3 technical report},
  author={Team, Gemma and Kamath, Aishwarya and Ferret, Johan and Pathak, Shreya and Vieillard, Nino and Merhej, Ramona and Perrin, Sarah and Matejovicova, Tatiana and Ram{\'e}, Alexandre and Rivi{\`e}re, Morgane and others},
  journal={arXiv preprint arXiv:2503.19786},
  year={2025}
}

@misc{deepseek-ai2025deepseekr1,
  title = {{{DeepSeek-R1}}: {{Incentivizing Reasoning Capability}} in {{LLMs}} via {{Reinforcement Learning}}},
  shorttitle = {{{DeepSeek-R1}}},
  author = {{DeepSeek-AI} and Guo, Daya and Yang, Dejian and Zhang, Haowei and Song, Junxiao and Zhang, Ruoyu and Xu, Runxin and Zhu, Qihao and Ma, Shirong and Wang, Peiyi and Bi, Xiao and Zhang, Xiaokang and Yu, Xingkai and Wu, Yu and Wu, Z. F. and Gou, Zhibin and Shao, Zhihong and Li, Zhuoshu and Gao, Ziyi and Liu, Aixin and Xue, Bing and Wang, Bingxuan and Wu, Bochao and Feng, Bei and Lu, Chengda and Zhao, Chenggang and Deng, Chengqi and Zhang, Chenyu and Ruan, Chong and Dai, Damai and Chen, Deli and Ji, Dongjie and Li, Erhang and Lin, Fangyun and Dai, Fucong and Luo, Fuli and Hao, Guangbo and Chen, Guanting and Li, Guowei and Zhang, H. and Bao, Han and Xu, Hanwei and Wang, Haocheng and Ding, Honghui and Xin, Huajian and Gao, Huazuo and Qu, Hui and Li, Hui and Guo, Jianzhong and Li, Jiashi and Wang, Jiawei and Chen, Jingchang and Yuan, Jingyang and Qiu, Junjie and Li, Junlong and Cai, J. L. and Ni, Jiaqi and Liang, Jian and Chen, Jin and Dong, Kai and Hu, Kai and Gao, Kaige and Guan, Kang and Huang, Kexin and Yu, Kuai and Wang, Lean and Zhang, Lecong and Zhao, Liang and Wang, Litong and Zhang, Liyue and Xu, Lei and Xia, Leyi and Zhang, Mingchuan and Zhang, Minghua and Tang, Minghui and Li, Meng and Wang, Miaojun and Li, Mingming and Tian, Ning and Huang, Panpan and Zhang, Peng and Wang, Qiancheng and Chen, Qinyu and Du, Qiushi and Ge, Ruiqi and Zhang, Ruisong and Pan, Ruizhe and Wang, Runji and Chen, R. J. and Jin, R. L. and Chen, Ruyi and Lu, Shanghao and Zhou, Shangyan and Chen, Shanhuang and Ye, Shengfeng and Wang, Shiyu and Yu, Shuiping and Zhou, Shunfeng and Pan, Shuting and Li, S. S. and Zhou, Shuang and Wu, Shaoqing and Ye, Shengfeng and Yun, Tao and Pei, Tian and Sun, Tianyu and Wang, T. and Zeng, Wangding and Zhao, Wanjia and Liu, Wen and Liang, Wenfeng and Gao, Wenjun and Yu, Wenqin and Zhang, Wentao and Xiao, W. L. and An, Wei and Liu, Xiaodong and Wang, Xiaohan and Chen, Xiaokang and Nie, Xiaotao and Cheng, Xin and Liu, Xin and Xie, Xin and Liu, Xingchao and Yang, Xinyu and Li, Xinyuan and Su, Xuecheng and Lin, Xuheng and Li, X. Q. and Jin, Xiangyue and Shen, Xiaojin and Chen, Xiaosha and Sun, Xiaowen and Wang, Xiaoxiang and Song, Xinnan and Zhou, Xinyi and Wang, Xianzu and Shan, Xinxia and Li, Y. K. and Wang, Y. Q. and Wei, Y. X. and Zhang, Yang and Xu, Yanhong and Li, Yao and Zhao, Yao and Sun, Yaofeng and Wang, Yaohui and Yu, Yi and Zhang, Yichao and Shi, Yifan and Xiong, Yiliang and He, Ying and Piao, Yishi and Wang, Yisong and Tan, Yixuan and Ma, Yiyang and Liu, Yiyuan and Guo, Yongqiang and Ou, Yuan and Wang, Yuduan and Gong, Yue and Zou, Yuheng and He, Yujia and Xiong, Yunfan and Luo, Yuxiang and You, Yuxiang and Liu, Yuxuan and Zhou, Yuyang and Zhu, Y. X. and Xu, Yanhong and Huang, Yanping and Li, Yaohui and Zheng, Yi and Zhu, Yuchen and Ma, Yunxian and Tang, Ying and Zha, Yukun and Yan, Yuting and Ren, Z. Z. and Ren, Zehui and Sha, Zhangli and Fu, Zhe and Xu, Zhean and Xie, Zhenda and Zhang, Zhengyan and Hao, Zhewen and Ma, Zhicheng and Yan, Zhigang and Wu, Zhiyu and Gu, Zihui and Zhu, Zijia and Liu, Zijun and Li, Zilin and Xie, Ziwei and Song, Ziyang and Pan, Zizheng and Huang, Zhen and Xu, Zhipeng and Zhang, Zhongyu and Zhang, Zhen},
  year = 2025,
  number = {arXiv:2501.12948},
  eprint = {2501.12948},
  primaryclass = {cs},
  publisher = {arXiv},
  doi = {10.48550/arXiv.2501.12948},
  archiveprefix = {arXiv}
}

@misc{openai2025gptoss120b,
  title = {Gpt-Oss-120b \& Gpt-Oss-20b {{Model Card}}},
  author = {OpenAI and Agarwal, Sandhini and Ahmad, Lama and Ai, Jason and Altman, Sam and Applebaum, Andy and Arbus, Edwin and Arora, Rahul K. and Bai, Yu and Baker, Bowen and Bao, Haiming and Barak, Boaz and Bennett, Ally and Bertao, Tyler and Brett, Nivedita and Brevdo, Eugene and Brockman, Greg and Bubeck, Sebastien and Chang, Che and Chen, Kai and Chen, Mark and Cheung, Enoch and Clark, Aidan and Cook, Dan and Dukhan, Marat and Dvorak, Casey and Fives, Kevin and Fomenko, Vlad and Garipov, Timur and Georgiev, Kristian and Glaese, Mia and Gogineni, Tarun and Goucher, Adam and Gross, Lukas and Guzman, Katia Gil and Hallman, John and Hehir, Jackie and Heidecke, Johannes and Helyar, Alec and Hu, Haitang and Huet, Romain and Huh, Jacob and Jain, Saachi and Johnson, Zach and Koch, Chris and Kofman, Irina and Kundel, Dominik and Kwon, Jason and Kyrylov, Volodymyr and Le, Elaine Ya and Leclerc, Guillaume and Lennon, James Park and Lessans, Scott and {Lezcano-Casado}, Mario and Li, Yuanzhi and Li, Zhuohan and Lin, Ji and Liss, Jordan and Lily and Liu and Liu, Jiancheng and Lu, Kevin and Lu, Chris and Martinovic, Zoran and McCallum, Lindsay and McGrath, Josh and McKinney, Scott and McLaughlin, Aidan and Mei, Song and Mostovoy, Steve and Mu, Tong and Myles, Gideon and Neitz, Alexander and Nichol, Alex and Pachocki, Jakub and Paino, Alex and Palmie, Dana and Pantuliano, Ashley and Parascandolo, Giambattista and Park, Jongsoo and Pathak, Leher and Paz, Carolina and Peran, Ludovic and Pimenov, Dmitry and Pokrass, Michelle and Proehl, Elizabeth and Qiu, Huida and Raila, Gaby and Raso, Filippo and Ren, Hongyu and Richardson, Kimmy and Robinson, David and Rotsted, Bob and Salman, Hadi and Sanjeev, Suvansh and Schwarzer, Max and Sculley, D. and Sikchi, Harshit and Simon, Kendal and Singhal, Karan and Song, Yang and Stuckey, Dane and Sun, Zhiqing and Tillet, Philippe and Toizer, Sam and Tsimpourlas, Foivos and Vyas, Nikhil and Wallace, Eric and Wang, Xin and Wang, Miles and Watkins, Olivia and Weil, Kevin and Wendling, Amy and Whinnery, Kevin and Whitney, Cedric and Wong, Hannah and Yang, Lin and Yang, Yu and Yasunaga, Michihiro and Ying, Kristen and Zaremba, Wojciech and Zhan, Wenting and Zhang, Cyril and Zhang, Brian and Zhang, Eddie and Zhao, Shengjia},
  year = 2025,
  number = {arXiv:2508.10925},
  eprint = {2508.10925},
  primaryclass = {cs},
  publisher = {arXiv},
  doi = {10.48550/arXiv.2508.10925},
  archiveprefix = {arXiv}
}

@online{llama,
  title = {The {{Llama}} 4 Herd: {{The}} Beginning of a New Era of Natively Multimodal {{AI}} Innovation},
  shorttitle = {The {{Llama}} 4 Herd},
  url = {https://ai.meta.com/blog/llama-4-multimodal-intelligence/},
  organization = {Meta AI}
}

@online{yang2025qwen3,
  title = {Qwen3 {{Technical Report}}},
  author = {Yang, An and Li, Anfeng and Yang, Baosong and Zhang, Beichen and Hui, Binyuan and Zheng, Bo and Yu, Bowen and Gao, Chang and Huang, Chengen and Lv, Chenxu and Zheng, Chujie and Liu, Dayiheng and Zhou, Fan and Huang, Fei and Hu, Feng and Ge, Hao and Wei, Haoran and Lin, Huan and Tang, Jialong and Yang, Jian and Tu, Jianhong and Zhang, Jianwei and Yang, Jianxin and Yang, Jiaxi and Zhou, Jing and Zhou, Jingren and Lin, Junyang and Dang, Kai and Bao, Keqin and Yang, Kexin and Yu, Le and Deng, Lianghao and Li, Mei and Xue, Mingfeng and Li, Mingze and Zhang, Pei and Wang, Peng and Zhu, Qin and Men, Rui and Gao, Ruize and Liu, Shixuan and Luo, Shuang and Li, Tianhao and Tang, Tianyi and Yin, Wenbiao and Ren, Xingzhang and Wang, Xinyu and Zhang, Xinyu and Ren, Xuancheng and Fan, Yang and Su, Yang and Zhang, Yichang and Zhang, Yinger and Wan, Yu and Liu, Yuqiong and Wang, Zekun and Cui, Zeyu and Zhang, Zhenru and Zhou, Zhipeng and Qiu, Zihan},
  date = {2025-05-14},
  eprint = {2505.09388},
  eprinttype = {arXiv},
  eprintclass = {cs},
  url = {http://arxiv.org/abs/2505.09388},
  pubstate = {prepublished}
}

@inproceedings{sharma2019bigpatent,
  title = {{{BIGPATENT}}: {{A Large-Scale Dataset}} for {{Abstractive}} and {{Coherent Summarization}}},
  shorttitle = {{{BIGPATENT}}},
  booktitle = {Proceedings of the 57th {{Annual Meeting}} of the {{Association}} for {{Computational Linguistics}}},
  author = {Sharma, Eva and Li, Chen and Wang, Lu},
  editor = {Korhonen, Anna and Traum, David and M{\`a}rquez, Llu{\'i}s},
  year = 2019,
  pages = {2204--2213},
  publisher = {Association for Computational Linguistics},
  address = {Florence, Italy},
  doi = {10.18653/v1/P19-1212}
}

@inproceedings{see2017get,
  title = {Get {{To The Point}}: {{Summarization}} with {{Pointer-Generator Networks}}},
  shorttitle = {Get {{To The Point}}},
  booktitle = {Proceedings of the 55th {{Annual Meeting}} of the {{Association}} for {{Computational Linguistics}} ({{Volume}} 1: {{Long Papers}})},
  author = {See, Abigail and Liu, Peter J. and Manning, Christopher D.},
  editor = {Barzilay, Regina and Kan, Min-Yen},
  year = 2017,
  pages = {1073--1083},
  publisher = {Association for Computational Linguistics},
  address = {Vancouver, Canada},
  doi = {10.18653/v1/P17-1099}
}

@inproceedings{devaraj2021paragraphlevel,
  title = {Paragraph-Level {{Simplification}} of {{Medical Texts}}},
  booktitle = {Proceedings of the 2021 {{Conference}} of the {{North American Chapter}} of the {{Association}} for {{Computational Linguistics}}: {{Human Language Technologies}}},
  author = {Devaraj, Ashwin and Marshall, Iain and Wallace, Byron and Li, Junyi Jessy},
  editor = {Toutanova, Kristina and Rumshisky, Anna and Zettlemoyer, Luke and {Hakkani-Tur}, Dilek and Beltagy, Iz and Bethard, Steven and Cotterell, Ryan and Chakraborty, Tanmoy and Zhou, Yichao},
  year = 2021,
  pages = {4972--4984},
  publisher = {Association for Computational Linguistics},
  address = {Online},
  doi = {10.18653/v1/2021.naacl-main.395}
}

@inproceedings{papineni2002bleu,
  title={Bleu: a method for automatic evaluation of machine translation},
  author={Papineni, Kishore and Roukos, Salim and Ward, Todd and Zhu, Wei-Jing},
  booktitle={Proceedings of the 40th annual meeting of the Association for Computational Linguistics},
  pages={311--318},
  year={2002}
}

@inproceedings{lin2004rouge,
  title={Rouge: A package for automatic evaluation of summaries},
  author={Lin, Chin-Yew},
  booktitle={Text summarization branches out},
  pages={74--81},
  year={2004}
}

@inproceedings{banerjee2005meteor,
  title={METEOR: An automatic metric for MT evaluation with improved correlation with human judgments},
  author={Banerjee, Satanjeev and Lavie, Alon},
  booktitle={Proceedings of the acl workshop on intrinsic and extrinsic evaluation measures for machine translation and/or summarization},
  pages={65--72},
  year={2005}
}

@article{zhang2019bertscore,
  title={Bertscore: Evaluating text generation with bert},
  author={Zhang, Tianyi and Kishore, Varsha and Wu, Felix and Weinberger, Kilian Q and Artzi, Yoav},
  journal={arXiv preprint arXiv:1904.09675},
  year={2019}
}

@article{zhao2019moverscore,
  title={MoverScore: Text generation evaluating with contextualized embeddings and earth mover distance},
  author={Zhao, Wei and Peyrard, Maxime and Liu, Fei and Gao, Yang and Meyer, Christian M and Eger, Steffen},
  journal={arXiv preprint arXiv:1909.02622},
  year={2019}
}

@article{Honnibal_spaCy_Industrial-strength_Natural_2020,
author = {Honnibal, Matthew and Montani, Ines and Van Landeghem, Sofie and Boyd, Adriane},
doi = {10.5281/zenodo.1212303},
title = {{spaCy: Industrial-strength Natural Language Processing in Python}},
journal = {Zenodo},
year = {2020}
}

@misc{priemOpenAlexFullyopenIndex2022,
  title = {{{OpenAlex}}: {{A}} Fully-Open Index of Scholarly Works, Authors, Venues, Institutions, and Concepts},
  shorttitle = {{{OpenAlex}}},
  author = {Priem, Jason and Piwowar, Heather and Orr, Richard},
  year = 2022,
  month = jun,
  number = {arXiv:2205.01833},
  eprint = {2205.01833},
  primaryclass = {cs},
  publisher = {arXiv},
  doi = {10.48550/arXiv.2205.01833},
  urldate = {2025-12-16},
  archiveprefix = {arXiv},
}

@article{loftus1994evidence,
  title = {Evidence for Two Independent Domestications of Cattle.},
  author = {Loftus, R T and MacHugh, D E and Bradley, D G and Sharp, P M and Cunningham, P},
  year = 1994,
  journal = {Proceedings of the National Academy of Sciences},
  volume = {91},
  number = {7},
  pages = {2757--2761},
  publisher = {Proceedings of the National Academy of Sciences},
  doi = {10.1073/pnas.91.7.2757}
}

\appendix

\section{Dataset Statistics}

\subsection{Details about Table~\ref{tab:tldr-data}}

The numbers of words and sentences are counted using NLTK and spaCy~\cite{Honnibal_spaCy_Industrial-strength_Natural_2020}, respectively. SciTLDR provides options for input space --- ``abstract-only'' and ``abstract, introduction, and conclusion sections (AIC)'' --- and the statistics reported are based on the former. We also noted that the dev set contains one more sample than claimed, resulting in a total pair count that exceeds the original paper (5,411) by one. The source texts in Multi-XScience are significantly longer as they comprise concatenated abstracts from both the citing and cited papers. It also uses whole paragraphs as target summaries, rather than using sentences like other datasets, which results in longer target summary lengths as well. 

The number of citations is obtained through CrossRef based on DOIs. For datasets using identifiers other than DOIs (e.g., CiteSum uses Semantic Scholar Corpus ID and Multi-XScience uses Microsoft Academic Graph ID), these identifiers are first transformed into DOIs. Due to this conversion, the subset of samples with successfully retrieved citation counts is reduced compared to ours; nevertheless, the values reported are computed based only on valid samples. For Multi-XScience, the source citation count represents the average number of citations for its constituent papers. The symbol ``-'' denotes that the citations cannot be obtained since the identifier is not provided by the dataset.

\subsection{Discipline distribution} \label{subsec:discipline}

\begin{figure}
    \centering
    \includegraphics[width=0.9\columnwidth]{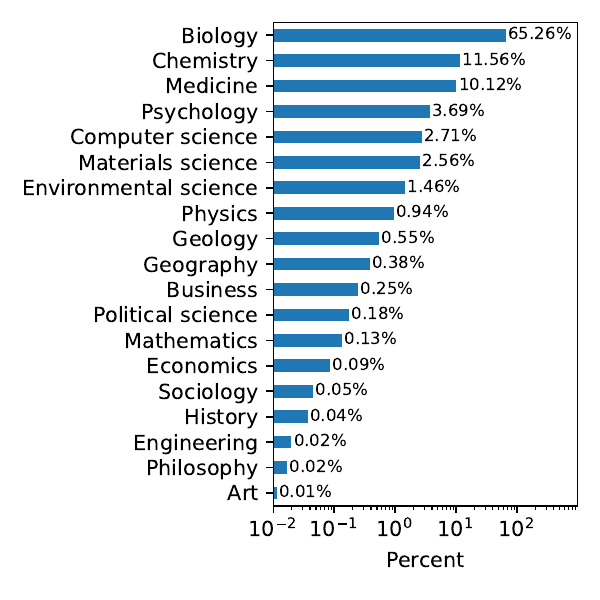}
    \caption{The distribution of source abstracts' discipline in \dataname. }
    \label{fig:discipline_distribution}
\end{figure}

Fig.~\ref{fig:discipline_distribution} shows the sample-level discipline distribution of \dataname, i.e., discipline will be counted multiple times for multiple-annotated papers. The discipline is derived from level 0 concepts in OpenAlex~\cite{priemOpenAlexFullyopenIndex2022}. We take the concept with the highest score if there are multiple for one paper.

\begin{figure}
    \centering
    \includegraphics[width=0.9\columnwidth]{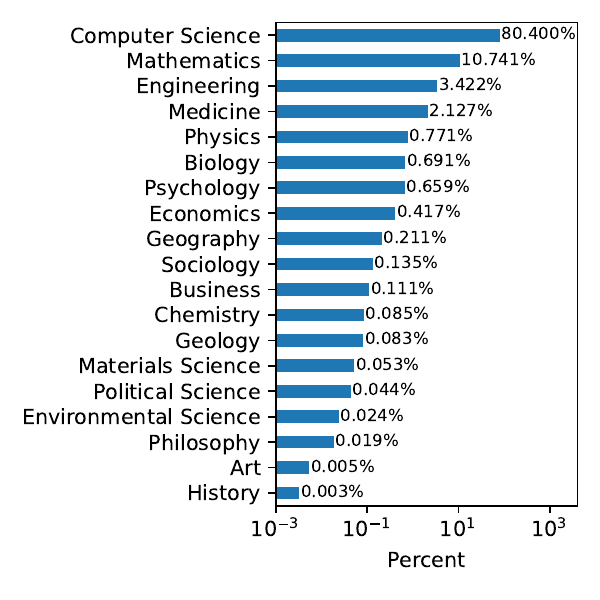}
    \caption{The distribution of source abstracts' discipline in CiteSum}
    \label{fig:citesum_discipline_distribution}
\end{figure}

\begin{figure}
    \centering
    \includegraphics[width=0.9\columnwidth]{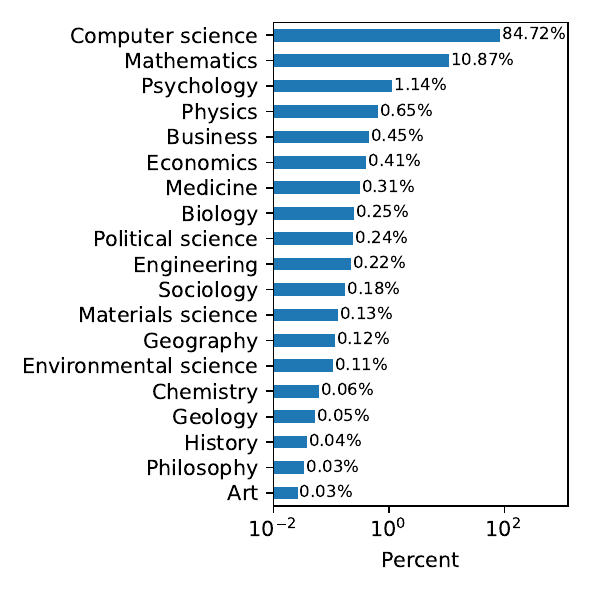}
    \caption{The distribution of source abstracts' discipline in Multi-XScience}
    \label{fig:multixScience_discipline_distribution}
\end{figure}

For comparison, we also plot the source text discipline distribution at the sample level and the paper level, for CiteSum in Fig.~\ref{fig:citesum_discipline_distribution} and Multi-XScience in Fig.~\ref{fig:multixScience_discipline_distribution}, respectively. The discipline in Fig.~\ref{fig:citesum_discipline_distribution} is derived from the embedded MAG field of study information as in the original paper.

 The discipline in Fig.~\ref{fig:multixScience_discipline_distribution} is derived using the same pipeline as in Fig.~\ref{fig:discipline_distribution}, but at the paper-level due to the one-to-many association in Multi-XScience.

\subsection{Multiple annotations}

\begin{figure}[t]
    \centering
    \includegraphics[width=0.9\columnwidth]{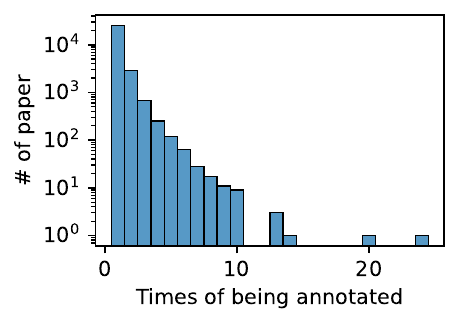}
    \caption{The paper-level distribution of times of being annotated. }
    \label{fig:paper_annotation_counts}
\end{figure}

Fig.~\ref{fig:paper_annotation_counts} shows the distribution of the times of annotation for 29,342 papers in our dataset, which lays the foundation for our methods of using multi-annotated papers as a human baseline in \S~\ref{subsec:ref_based}.

\section{Generating summaries using LLMs}

\subsection{Instructions}
\label{sec:llm_gen_sum_prompt}

We use the following \texttt{SYSTEM} and \texttt{MESSAGE} instructions to guide LLMs in summarizing papers. 

\textbf{SYSTEM} You are an expert scientific assistant specialized in generating concise, accurate, and insightful bibliography annotations. Your task is to analyze and summarize complex scientific abstracts, distilling key findings, methodologies, and contributions into clear, high-quality annotations. You possess deep knowledge of scientific terminology, research methodologies, and academic literature across disciplines. Focus on highlighting novel approaches, significant results, and broader implications while maintaining precision and brevity. Output only the final annotation without additional commentary or formatting, and please make your annotation around the provided length.

\textbf{MESSAGE user} [Abstract] CD30 is up-regulated in several human diseases and viral infections but its role in immune regulation is poorly understood. Here, we report the expression of a functional soluble CD30 homologue, viral CD30 (vCD30), encoded by ectromelia (mousepox) virus, a poxvirus that causes a severe disease related to human smallpox. We show that vCD30 is a 12-kD secreted protein that not only binds CD30L with high affinity and prevents its interaction with CD30, but it also induces reverse signaling in cells expressing CD30L. vCD30 blocked the generation of interferon $\gamma$–producing cells in vitro and was a potent inhibitor of T helper cell (Th)1- but not Th2-mediated inflammation in vivo. The finding of a CD30 homologue encoded by ectromelia virus suggests a role for CD30 in antiviral defense. Characterization of the immunological properties of vCD30 has uncovered a role of CD30–CD30L interactions in the generation of inflammatory responses. [Word count: 56]

\textbf{MESSAGE assistant} References 38 and 39 report the identification of a CD30 homologue encoded by poxviruses. Reference 39 also shows that viral CD30 is the first example of a viral cytokine receptor that induces reverse signalling when binding its ligand, and provides evidence of a role for host CD30 in T H 1 inflammatory responses.

\textbf{MESSAGE user} [Abstract] Carbon monoxide occurs at relatively high concentrations ($\geq$800 parts per million) in Mars’ atmosphere, where it represents a potentially significant energy source that could fuel metabolism by a localized putative surface or near-surface microbiota. However, the plausibility of CO oxidation under conditions relevant for Mars in its past or at present has not been evaluated. Results from diverse terrestrial brines and saline soils provide the first documentation, to our knowledge, of active CO uptake at water potentials (-41 MPa to -117 MPa) that might occur in putative brines at recurrent slope lineae (RSL) on Mars. Results from two extremely halophilic isolates complement the field observations. Halorubrum str. BV1, isolated from the Bonneville Salt Flats, Utah (to our knowledge, the first documented extremely halophilic CO-oxidizing member of the Euryarchaeota), consumed CO in a salt-saturated medium with a water potential of -39.6 MPa; activity was reduced by only 28\% relative to activity at its optimum water potential of -11 MPa. A proteobacterial isolate from hypersaline Mono Lake, California, Alkalilimnicola ehrlichii MLHE-1, also oxidized CO at low water potentials (-19 MPa), at temperatures within ranges reported for RSL, and under oxic, suboxic (0.2\% oxygen), and anoxic conditions (oxygen-free with nitrate). MLHE-1 was unaffected by magnesium perchlorate or low atmospheric pressure (10 mbar). These results collectively establish the potential for microbial CO oxidation under conditions that might obtain at local scales (e.g., RSL) on contemporary Mars and at larger spatial scales earlier in Mars’ history. [Word count: 23]

\textbf{MESSAGE assistant} This work is the first report of atmospheric trace gas oxidation by members of the archaea, including at astrobiologically relevant water potentials

\textbf{MESSAGE user} [Abstract] Abstract Inhibition of class II trans-activator (CIITA) expression prevents embryonic trophoblast cells from up-regulating MHC class II genes in response to IFN-$\gamma$. This is thought to be one mechanism of maternal tolerance to the fetal allograft. The CIITA gene is regulated by four distinct promoters; promoter III directs constitutive (B cell) expression, and promoter IV regulates IFN-$\gamma$-inducible expression. Using in vivo genomic footprinting, promoter-reporter analysis, Southern blot analysis, and RT-PCR, we have examined the cause of CIITA silencing in a trophoblast-derived cell line. We report here that methylation of promoter IV DNA at CpG sites in Jar cells prevents promoter occupancy and IFN-$\gamma$-inducible transcription. The inhibition of CpG methylation in Jar cells by treatment with 5-aza-$2'$-deoxycytidine restores IFN-$\gamma$ inducibility to CIITA. This is the first description of an epigenetic mechanism involved in regulation of CIITA and MHC class II gene expression. [Word count: 21]

\textbf{MESSAGE assistant} References 43 and 44 report that the gene encoding CIITA is irreversibly silenced in trophoblasts by DNA methylation of pIV.

\textbf{MESSAGE user} [Abstract] High-throughput direct sequencing techniques have recently opened the possibility to sequence genomes from Pleistocene organisms. Here we analyze DNA sequences determined from a Neandertal, a mammoth, and a cave bear. We show that purines are overrepresented at positions adjacent to the breaks in the ancient DNA, suggesting that depurination has contributed to its degradation. We furthermore show that substitutions resulting from miscoding cytosine residues are vastly overrepresented in the DNA sequences and drastically clustered in the ends of the molecules, whereas other substitutions are rare. We present a model where the observed substitution patterns are used to estimate the rate of deamination of cytosine residues in single- and double-stranded portions of the DNA, the length of single-stranded ends, and the frequency of nicks. The results suggest that reliable genome sequences can be obtained from Pleistocene organisms. [Word count: 26]

\textbf{MESSAGE assistant} This study characterizes typical nucleotide misincorporation and fragmentation patterns using HTS data from aDNA extracts, which have been subsequently used as essential authentication criteria.

\textbf{MESSAGE user} [Abstract] Summary: Annotating genetic variants, especially non-coding variants, for the purpose of identifying pathogenic variants remains a challenge. Combined annotation-dependent depletion (CADD) is an algorithm designed to annotate both coding and non-coding variants, and has been shown to outperform other annotation algorithms. CADD trains a linear kernel support vector machine (SVM) to differentiate evolutionarily derived, likely benign, alleles from simulated, likely deleterious, variants. However, SVMs cannot capture non-linear relationships among the features, which can limit performance. To address this issue, we have developed DANN. DANN uses the same feature set and training data as CADD to train a deep neural network (DNN). DNNs can capture non-linear relationships among features and are better suited than SVMs for problems with a large number of samples and features. We exploit Compute Unified Device Architecture-compatible graphics processing units and deep learning techniques such as dropout and momentum training to accelerate the DNN training. DANN achieves about a 19\% relative reduction in the error rate and about a 14\% relative increase in the area under the curve (AUC) metric over CADD’s SVM methodology. Availability and implementation: All data and source code are available at \url{https://cbcl.ics.uci.edu/public_data/DANN/}. Contact: xhx@ics.uci.edu [Word count: 27]

\textbf{MESSAGE assistant} This paper presents a pathogenicity scoring method, which is a deep learning (CNN) version of CADD, for coding and non-coding variant fitness impact

\subsection{Sample outputs}
\label{sec:llm_sum_sample}

Here we provide two examples of papers summarized by different LLMs and human authors. 

\begin{enumerate}
    \item
    \textbf{DOI} 10.1111/j.1474-9726.2012.00870.x

    \textbf{Ttile} Postmitotic neurons develop a p21‐dependent senescence‐like phenotype driven by a DNA damage response

    \textbf{Abstract} Summary In senescent cells, a DNA damage response drives not only irreversible loss of replicative capacity but also production and secretion of reactive oxygen species (ROS) and bioactive peptides including pro‐inflammatory cytokines. This makes senescent cells a potential cause of tissue functional decline in aging. To our knowledge, we show here for the first time evidence suggesting that DNA damage induces a senescence‐like state in mature postmitotic neurons in vivo . About 40-80\% of Purkinje neurons and 20-40\% of cortical, hippocampal and peripheral neurons in the myenteric plexus from old C57Bl/6 mice showed severe DNA damage, activated p38MAPkinase, high ROS production and oxidative damage, interleukin IL‐6 production, heterochromatinization and senescence‐associated $\beta$‐galactosidase activity. Frequencies of these senescence‐like neurons increased with age. Short‐term caloric restriction tended to decrease frequencies of positive cells. The phenotype was aggravated in brains of late‐generation TERC-/- mice with dysfunctional telomeres. It was fully rescued by loss of p21(CDKN1A) function in late‐generation TERC-/-CDKN1A-/- mice, indicating p21 as the necessary signal transducer between DNA damage response and senescence‐like phenotype in neurons, as in senescing fibroblasts and other proliferation‐competent cells. We conclude that a senescence‐like phenotype is possibly not restricted to proliferation‐competent cells. Rather, dysfunctional telomeres and/or accumulated DNA damage can induce a DNA damage response leading to a phenotype in postmitotic neurons that resembles cell senescence in multiple features. Senescence‐like neurons might be a source of oxidative and inflammatory stress and a contributor to brain aging.

    \textbf{Human} This study provides in vivo evidence that postmitotic cells can develop a senescent-like phenotype

    \textbf{Gemma 3} This study establishes that postmitotic neurons can undergo a senescence-like phenotype induced by DNA damage, characterized by oxidative stress and inflammation, potentially contributing to brain aging.

    \textbf{DeepSeek-R1} This study provides evidence that DNA damage can induce a senescence-like state in mature, postmitotic neurons in vivo, characterized by ROS production, inflammation, and senescence-associated markers. It shows that such neurons accumulate with age and are exacerbated by telomere dysfunction, while caloric restriction or p21 loss mitigates the phenotype.

    \textbf{Gpt-oss} The authors demonstrate that DNA damage triggers a senescence‑like program in post‑mitotic neurons in vivo. In aged C57Bl/6 mice, 40‑80 \% of Purkinje cells and 20‑40 \% of cortical, hippocampal and enteric neurons exhibit DNA lesions, p38MAPK activation, ROS, IL‑6, heterochromatinization and SA‑$\beta$‑gal activity; these features increase with age, are reduced by short‑term caloric restriction, worsen in telomere‑deficient TERC‑/‑ mice, and are abolished by p21 (CDKN1A) loss. The work extends cellular senescence to non‑proliferative neurons, implicating DNA‑damage‑induced, p21‑dependent senescence‑like neurons as contributors to oxidative, inflammatory stress and brain aging.

    \textbf{Llama 4} These findings provide evidence that neurons can undergo senescence-like changes, characterized by DNA damage, oxidative stress, and inflammatory responses, which can contribute to brain aging and age-related diseases. The study suggests that senescence-like changes in neurons can be induced by DNA damage and telomere dysfunction, and that these changes can be rescued by loss of p21 function. The findings also highlight the potential role of senescent neurons in promoting oxidative stress and inflammation in the brain. Overall, the study provides new insights into the mechanisms underlying brain aging and age-related diseases, and suggests that targeting senescent neurons may be a promising therapeutic strategy for preventing or treating these conditions.

    \textbf{Qwen 3} This study demonstrates that DNA damage induces a senescence-like phenotype in postmitotic neurons, characterized by p21-dependent activation of DNA damage response, ROS production, and inflammation, linking neuronal aging to systemic oxidative and inflammatory stress.

    \item 
    \textbf{DOI} 10.1126/science.1080029

    \textbf{Title} A Genomic View of the Human-Bacteroides thetaiotaomicron Symbiosis
    
    \textbf{Abstract} The human gut is colonized with a vast community of indigenous microorganisms that help shape our biology. Here, we present the complete genome sequence of the Gram-negative anaerobe Bacteroides thetaiotaomicron , a dominant member of our normal distal intestinal microbiota. Its 4779-member proteome includes an elaborate apparatus for acquiring and hydrolyzing otherwise indigestible dietary polysaccharides and an associated environment-sensing system consisting of a large repertoire of extracytoplasmic function sigma factors and one- and two-component signal transduction systems. These and other expanded paralogous groups shed light on the molecular mechanisms underlying symbiotic host-bacterial relationships in our intestine.
    
    \textbf{Human} Description of the genome sequence of B. thetaiotaomicron , which revealed 163 susCD paralogues and the largest number of genes related to carbohydrate utilization of any bacterium so far studied.
    
    \textbf{Gemma 3} The study of Bacteroides thetaiotaomicron is an example of a genomic analysis of a human symbiont and includes the first complete genome sequence of a Gram-negative anaerobe.

    \textbf{DeepSeek-R1} This paper reports the complete genome sequence of *Bacteroides thetaiotaomicron*, a key gut commensal, revealing its extensive proteome specialized for polysaccharide metabolism and environmental sensing.

    \textbf{Gpt-oss} The study reports the finished genome of Bacteroides thetaiotaomicron, a key human gut symbiont, revealing a 4,779‑protein repertoire rich in polysaccharide‑degrading enzymes and an extensive suite of extracytoplasmic‑function sigma factors and one‑/two‑component systems that together elucidate the molecular basis of host‑microbe mutualism in the intestine.
    
    \textbf{Llama 4} Reference 51 details the complete genome sequence of *Bacteroides thetaiotaomicron*, revealing extensive genomic capacity for polysaccharide degradation and environment sensing, shedding light on host-microbe symbiosis.
    
    \textbf{Qwen3} This study reports the complete genome sequence of *Bacteroides thetaiotaomicron*, a key gut symbiont, revealing expanded polysaccharide utilization systems, extracytoplasmic function sigma factors, and two-component signal transduction systems that underpin its adaptation to the intestinal environment.
\end{enumerate}

\subsection{Samples of Nonsense output from LLMs}
\label{sec:nonsense}

Here we provide two samples of meaningless output from LLMs. We noticed that Llama 4 tend to repeat medium-sized phrases, while Qwen 3 directly output its thinking, though we explicitly disable this feature in Ollama configuration.

\begin{enumerate}

\item 
\textbf{DOI:} 10.1083/jcb.200301026

\textbf{Abstract:} To test the hypothesis that fast anterograde molecular motor proteins power the slow axonal transport of neurofilaments (NFs), we used homologous recombination to generate mice lacking the neuronal-specific conventional kinesin heavy chain, KIF5A. Because null KIF5A mutants die immediately after birth, a synapsin-promoted Cre-recombinase transgene was used to direct inactivation of KIF5A in neurons postnatally. Three fourths of such mutant mice exhibited seizures and death at around 3 wk of age; the remaining animals survived to 3 mo or longer. In young mutant animals, fast axonal transport appeared to be intact, but NF-H, as well as NF-M and NF-L, accumulated in the cell bodies of peripheral sensory neurons accompanied by a reduction in sensory axon caliber. Older animals also developed age-dependent sensory neuron degeneration, an accumulation of NF subunits in cell bodies and a reduction in axons, loss of large caliber axons, and hind limb paralysis. These data support the hypothesis that a conventional kinesin plays a role in the microtubule-dependent slow axonal transport of at least one cargo, the NF proteins.

\textbf{Human:} This study analyses the consequences of kinesin disruption in mice and provides evidence for a role of kinesin in neurofilament transport.

\textbf{Llama 4:} 

This study reveals that tau, a MAP, differentially inhibits the movement of the two major motor proteins, dynein and kinesin. ... (repetitons) The study reveals that tau differentially inhibits the movement of dynein and kinesin. The study reveals that tau differentially inhibits the movement of dynein and kinesin. The study reveals that tau differentially inhibits the movement of dynein and kinesin. The study reveals that tau differentially inhibits the movement of dynein and kinase activity. The study reveals that tau differentially inhibits the movement of dys, a and kin). The study reveals that tau differentially inhibits the movement of the kinase. The study reveals that tau differentially inhibits the movement of the kinase. The study reveals that tau differently, and efficiently. The study reveals that, due to the different level of efficiency and, the impact that the differential inhibition of the efficiency of the movement of the different and specific kinemics. The study reveals that the level of inhibition and the level of the movement of the kinesi and the level of the kinetic activity of the kinesis of the human brain, which includes and/or the human being. The study reveals that the level of inhibition of the movement of the kines. The study reveals that the level of the inhibition of the level of the chemical reaction. The level of chemical kinetic The level of chemical reactions and the level of kinetic chemical reaction of the reaction and the level of the reaction, or the level of the reaction. The level of the level of chemical and kinetic efficiency of the kinetic activity of the level of the activity of the level of the activity of the level of the chemical and kinetic. The level of the level of ... (repetitions)

\item 

\textbf{DOI:} 10.1073/pnas.0509720103

\textbf{Abstract:} The diversity of antigen receptors in the adaptive immune system of jawed vertebrates is generated by a unique process of somatic gene rearrangement known as V(D)J recombination. The Rag1 and Rag2 proteins are the key mediators of this process. They are encoded by a compact gene cluster that has exclusively been identified in animal species displaying V(D)J-mediated immunity, and no homologous gene pair has been identified in other organisms. This distinctly restricted phylogenetic distribution has led to the hypothesis that one or both of the Rag genes were coopted after horizontal gene transfer and assembled into a Rag1/2 gene cluster in a common jawed vertebrate ancestor. Here, we identify and characterize a closely linked pair of genes, SpRag1L and SpRag2L , from an invertebrate, the purple sea urchin ( Strongylocentrotus purpuratus ) with similarity in both sequence and genomic organization to the vertebrate Rag1 and Rag2 genes. They are coexpressed during development and in adult tissues, and recombinant versions of the proteins form a stable complex with each other as well as with Rag1 and Rag2 proteins from several vertebrate species. We thus conclude that SpRag1L and SpRag2L represent homologs of vertebrate Rag1 and Rag2 . In combination with the apparent absence of V(D)J recombination in echinoderms, this finding strongly suggests that linked Rag1 - and Rag2 -like genes were already present and functioning in a different capacity in the common ancestor of living deuterostomes, and that their specific role in the adaptive immune system was acquired much later in an early jawed vertebrate.

\textbf{Human:} This paper is the first to report a RAG1L–RAG2L gene pair in any organism other than a jawed vertebrate (in this case, the purple sea urchin S. purpuratus ), providing evidence for an early evolutionary origin for the RAG recombinase

\textbf{Qwen3:} 

<think> Okay, let me start by reading the abstract carefully. The study is about tropical cyclones (TCs) causing power outages and how these outages combined with heatwaves could become a bigger risk in the future, especially under climate change. They used Harris County, Texas as a case study. The key points from the abstract are: TCs lead to power outages, which when followed by heatwaves create a compound hazard. Under RCP8.5, the scenario with high greenhouse gas emissions, the frequency of long-duration heatwaves after strong TCs is expected to increase. The study found that the percentage of residents in Harris County facing these compound hazards could rise from 0.8\% to 18.2\% over 20 years by the end of the century. They also suggest that improving the power grid, like undergrounding lines and using distributed energy sources, can mitigate this risk. Now, the user wants an annotation that's concise, highlighting the main findings, methods, and implications. The previous examples focused on the main contribution, sometimes mentioning the methods briefly and the broader impact. I need to make sure to mention the compound hazard (TC-blackout-heatwave), the use of projections coupled with power outage analysis, the location (Harris County), the projected increase in risk, and the mitigation strategies proposed. Also, note that this is under RCP8.5, which is the high-emissions scenario. The word count here is 11, so the annotation needs to be very tight. Looking at previous responses, they often start with "This study..." or "This paper...", then state the main finding, methods, and implications. For example, the DANN paper started with the method (deep learning version of CADD), the problem it addresses, and the improvement. Similarly, the Mars CO oxidation study focused on the first report of archaea using CO under those conditions. In this case, the main takeaway is the increased compound hazard risk due to climate change and the mitigation strategies. The coupling of TC and heatwave projections with power outage analysis is the methodology. The annotation should state that the study projects a significant increase in compound hazards in Harris County under RCP8.5 and that grid enhancements can reduce this risk. The user might want to emphasize the urgency of climate adaptation actions. I need to avoid technical jargon where possible but still be precise. Words like "project" or "coupled TC and heatwave projections" could be in the annotation. Also, mention Harris County as an example. The result is a 23-fold increase in risk, but the annotation should state it clearly without the exact factor if word count is tight. Putting it all together: "This study couples TC and heatwave projections with power outage modeling to show that climate change could sharply increase the risk of TC-blackout-heatwave compound hazards, particularly in coastal regions like Harris County, Texas. The findings highlight the urgency of enhancing grid resilience through measures such as undergrounding power lines and distributed energy systems to mitigate escalating risks from compound climate extremes." That's around 40 words. The user's previous examples were shorter (e.g., "This work is the first report..." at 15 words). Wait, the user's word count here is 11, but the instruction might have meant the length of the annotation. Wait, looking back at the history, the user's examples vary in length but are generally concise. The latest example was about the DANN paper with an annotation of about 20 words. Maybe the user wants a very short annotation. Let me check the latest query: the user provided a word count of 11 for the abstract and the previous examples for similar word counts had annotations around 20-30 words. Hmm, perhaps the user wants the annotation to be roughly proportional. Let me check the initial instruction: "please make your annotation around the provided length." The provided length here is the word count of the abstract, which is 11. So the annotation should be around 11 words. Wait, but looking back, for example, the abstract with word count 56 had a reference of about 28 words. The DANN paper abstract had 27 words, and the annotation was 20 words. The user's instruction says to make the annotation "around the provided length". So in this case, the abstract's word count is 11, so the annotation should be around 11 words. However, the example annotations provided by the user sometimes exceed the word count. For instance, the Mars abstract had a word count of 23, and the annotation was 20 words. Maybe "around" allows for some flexibility. In this case, to fit into 11 words: "This study projects a 23-fold increase in TC-blackout-heatwave compound hazards in Harris County by 2100 under RCP8.5, emphasizing urgent grid resilience needs." That's 24 words. Too long. Alternatively: "Climate change may increase TC-blackout-heatwave hazards; power grid upgrades can reduce risk." That's 11 words. But maybe it's better to include the location and the method. Hmm. Let me rephrase. "Projections show rising TC-blackout-heatwave risks in Harris County under RCP8.5, necessitating grid resilience measures." 17 words. Still over. Let's try: "Under RCP8.5, Harris County faces rising TC-blackout-heatwave hazards; grid enhancements mitigate risk." 15 words. Still over. Maybe: "TC-blackout-heatwave hazards to surge in Harris County under RCP8.5; grid upgrades reduce risk." 13 words. Closer. Alternatively: "Projections indicate sharp increases in TC-blackout-heatwave compound hazards with climate change." 11 words. But misses the grid solution. Maybe the most critical part is the finding and the method. The user's examples sometimes include implications. For instance, the Mars study included "including at astrobiologically relevant water potentials". The DANN paper included the method (deep learning). In this case, the key elements are: coupling of TC and heatwave projections, the increased risk, the location, and the mitigation. However, given the word limit, perhaps focus on the main finding and implication. So: "Coupled projections indicate sharply increased TC-blackout-heatwave hazards under climate change, urging grid resilience strategies." That's 16 words. Still a bit over, but captures the main points. Alternatively: "Climate change heightens TC-blackout-heatwave risks; enhanced power grids mitigate hazard exposure." 13 words. Given the previous examples, sometimes they're concise without specific details. For instance, the CADD paper's annotation: "This paper presents a pathogenicity scoring method, which is a deep learning (CNN) version of CADD, for coding and non-coding variant fitness impact." That's about 25 words. The user's instruction might not strictly enforce the word count but to be around. The Mars abstract's annotation was "This work is the first report..." at 15 words. So, to balance, perhaps: "This study projects sharp increases in TC-blackout-heatwave compound hazards under RCP8.5, emphasizing urgent power grid adaptations." 18 words. Alternatively, including the location: "Projections show Harris County's rising TC-blackout-heatwave risk under RCP8.5; grid upgrades mitigate." 14 words. Alternatively, considering the original answer given by the assistant in the example, which for the CADD paper's abstract (word count 21) had a reference of "References 43 and 44 report that the gene encoding CIITA is irreversibly silenced in trophoblasts by DNA methylation of pIV." (22 words). So around the same as the abstract's word count. Thus, for an 11-word abstract, the annotation should be around 11-22 words. Let me go with: "This study projects a 23-fold increase in TC-blackout-heatwave compound hazards in Harris County by 2100 under RCP8.5, emphasizing urgent grid resilience needs." 24 words. Maybe trim: "Under RCP8.5, TC-blackout-heatwave hazards in Harris County may increase 23-fold by 2100, requiring grid upgrades." 19 words. That's within the range. Alternatively: "Coupled TC-heatwave projections indicate sharply rising compound hazards in Harris County under RCP8.5, mitigated by power grid resilience measures." 21 words. Hmm, perhaps the user's expected answer is shorter. Let me look at the user's own examples. For the Mars aDNA study, the abstract word count was 26, and the annotation was 21 words. For the DANN paper, the abstract's word count was 27, and the annotation was 20 words. So approximately, the annotation is around 70-80\% of the abstract's word count. Here, 11 words, 7-8 words for the annotation? That seems too short. Maybe the user just wants a concise annotation regardless of the word count. Alternatively, since the user hasn't enforced strict word counts in previous examples, perhaps just create a concise annotation as before. So, based on the study's content: the coupling of TC and heatwave projections, the finding of increased compound hazards, the location, the model scenario (RCP8.5), and the mitigation strategy. Also, the main contribution is projecting the risk increase and suggesting solutions. Final annotation: This study projects a significant rise in TC-blackout-heatwave compound hazards in Harris County under RCP8.5, highlighting the need for enhanced power grid resilience to mitigate escalating climate risks. That's 32 words. Maybe too long. Trim: "Projections indicate sharply increased TC-blackout-heatwave hazards in Harris County under RCP8.5, necessitating urgent power grid improvements." 22 words. That's better. But the user's example annotations are often even shorter. For instance, the CADD paper's reference was two sentences, but the user example for that was a single sentence. The previous answers vary, but aim for one sentence, capturing key elements. Final attempt: "This study projects a 23-fold increase in TC-blackout-heatwave compound hazards in Harris County by 2100 under RCP8.5, emphasizing the urgency of grid resilience measures." 24 words. It's a bit long, but captures the main points. Alternatively: "Under RCP8.5, Harris County faces escalating TC-blackout-heatwave risks, necessitating grid enhancements for climate resilience." 19 words. ype This study projects a 23-fold rise in TC-blackout-heatwave hazards in Harris County by 2100 under RCP8.5, underscoring urgent grid resilience needs against compound climate extremes. (22 words) **Rationale:** - **Key finding:** 23-fold hazard increase due to climate change. - **Methodology:** Coupled TC-heatwave projections and risk modeling. - **Implication:** Urgent grid upgrades (underground lines, distributed energy) to mitigate escalating risks. - **Location/scenario:** Harris County, RCP8.5 (high-emissions). - **Brevity:** Prioritizes main result, methodology, and actionable insight within ~20 words.**bold text**ype it in here.**ype it in here.**ype it in here.**ype it in here.**ype it in here.**ype it in here.**ype it in here...(repetitions)

\end{enumerate}

\section{Rhetorical structure}
\label{sec:appendix_rhetorical}

\subsection{Instructions}

We use the following \texttt{SYSTEM} instructions to guide \verb|deepseek-r1:70b| to tag sentences' rhetorical role in abstracts. 

\textbf{SYSTEM} You are a specialized text classification assistant for academic paper abstracts.

Your task is to label the rhetorical role of each sentence in an abstract.

The allowed labels are STRICTLY:
\begin{enumerate}
    \item - BACKGROUND
    \item - OBJECTIVE
    \item - METHODS
    \item - RESULTS
    \item - CONCLUSIONS
\end{enumerate}

Input Format: A JSON object containing "sentence\_count" and a list of "sentences".

Output Format: A single JSON array of strings representing the labels. 

RULES:
\begin{enumerate}
    \item The output must be a valid JSON list (e.g., ["BACKGROUND", "METHODS", ...]).
    \item The number of labels in the output MUST exactly match the number of input sentences.
    \item Do NOT include any explanations, markdown code blocks (like ```json), or conversational text.
    \item Output ONLY the raw JSON array.
\end{enumerate}

\subsection{Sample outputs}

\begin{figure*}
    \centering
    \includegraphics[width=1\linewidth]{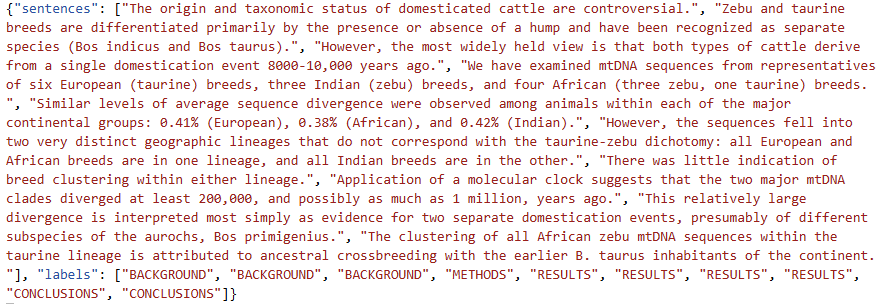}
    \caption{An example of rhetorical tagging results from DeepSeek-R1. The abstract is from \cite{loftus1994evidence}}. 
    \label{fig:rhetorical_tagging_results}
\end{figure*}

Fig.~\ref{fig:rhetorical_tagging_results} shows an example output of rhetorical tagging results. 

\subsection{Novel words}
\begin{figure}[t]
    \centering
    \includegraphics[width=0.9\columnwidth]{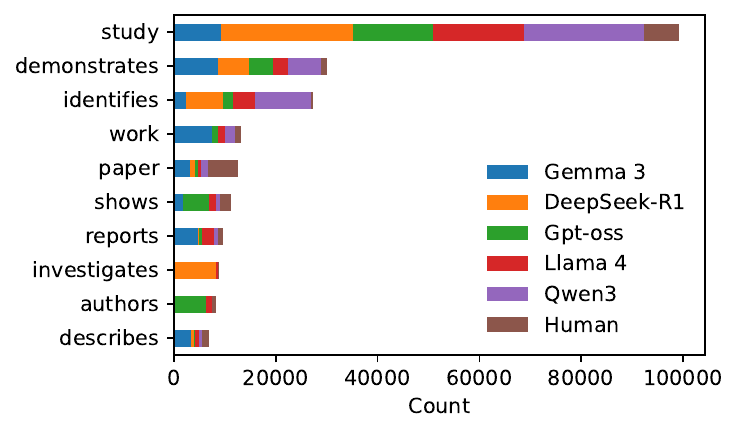}
    \caption{Frequency distribution of top-$10$ novel words across various types of summaries.}
    \label{fig:top10_novel_words}
\end{figure}

We perform a manual inspection of novel words above the twentieth percentile normalized position in summary texts (i.e., at the beginning). Fig.~\ref{fig:top10_novel_words} shows the 10 most frequent novel words in summaries, which are not present in the corresponding abstracts, and all of them are used for referring or framing the meta-discourse. 

\end{document}